\documentclass{aims} 
\usepackage{amsmath}
\usepackage{paralist}
\usepackage[misc]{ifsym}
\usepackage{epsfig} 
\usepackage{epstopdf} 
\usepackage[colorlinks=true]{hyperref}
\hypersetup{urlcolor=blue, citecolor=red}
\allowdisplaybreaks

\def\currentvolume{X}
 \def\currentissue{X}
  \def\currentyear{200X}
   \def\currentmonth{XX}
    \def\ppages{X-XX}

\theoremstyle{definition}

\usepackage{graphicx} 
\usepackage{xcolor}
\usepackage{enumitem}
\usepackage{dirtytalk}
\usepackage{tikz}
\usepackage{bbold}
\usepackage{sidecap}
\usepackage{color, colortbl}
\usepackage{float}
\usepackage{amssymb,mathtools}
\usepackage[nobysame]{amsrefs}
\usepackage[normalem]{ulem}
\usepackage{algorithm, algpseudocode, amsmath}
\usepackage{multirow}
\usepackage{tabularx}
\usepackage{tabularray}
\usepackage{rotating}
\usepackage{array}
\usetikzlibrary{arrows,calc,fit}
\usepackage{forest}
\usepackage{caption}
\usepackage{subcaption}

\DeclareMathOperator*{\Cov}{Cov}
\definecolor{MatplotOrange}{HTML}{ff7f0e}
\definecolor{MatplotBlue}{HTML}{1f77b4}
\definecolor{MatplotGreen}{HTML}{2ca02c}
\definecolor{MatplotRed}{HTML}{d62728}
\usetikzlibrary{arrows.meta}
\usetikzlibrary{matrix}
\tikzset{every picture/.style={/utils/exec={\sffamily}}}
\usepackage{lipsum} 
\usepackage{booktabs}
\usepackage{hhline}
\usepackage{url}
\usepackage{listings}

\usepackage{xcolor}

\definecolor{codegreen}{rgb}{0,0.6,0}
\definecolor{codegray}{rgb}{0.5,0.5,0.5}
\definecolor{codepurple}{rgb}{0.58,0,0.82}
\definecolor{backcolour}{rgb}{0.95,0.95,0.92}

\lstdefinestyle{mystyle}{
    backgroundcolor=\color{backcolour},   
    commentstyle=\color{codegreen},
    keywordstyle=\color{magenta},
    numberstyle=\tiny\color{codegray},
    stringstyle=\color{codepurple},
    basicstyle=\ttfamily\footnotesize,
    breakatwhitespace=false,         
    breaklines=true,                 
    captionpos=b,                    
    keepspaces=true,                 
    numbers=left,                    
    numbersep=5pt,                  
    showspaces=false,                
    showstringspaces=false,
    showtabs=false,                  
    tabsize=2
}

\numberwithin{equation}{section}

\newcommand{\rex}[1]{\textcolor{black}{#1}}
\newcommand{\rey}[1]{\textcolor{black}{#1}}
\newcommand{\rez}[1]{{\color{black}{#1}}}
\newcommand{\ergys}[1]{{\color{black}{#1}}}

\definecolor{mygray}{gray}{0.95}

\title[Supervised TSC for Anomaly Detection in Subsea Engineering]
{Supervised Time Series Classification for Anomaly Detection in Subsea Engineering} 

\author[E. Çokaj, H. S. Gustad, A. Leone, P. T. Moe and L. Moldestad]{}



\thanks{$^*$Corresponding author: Halvor Snersrud Gustad}

\begin{document}
\maketitle

\centerline{
\scshape
Ergys Çokaj$^{{\href{mailto:ergys.cokaj@ntnu.no}{\textrm{\Letter}}}1}$, 
Halvor Snersrud Gustad$^{{\href{mailto:halvorsnersrud.gustad@technipfmc.com}{\textrm{\Letter}}}*1,2}$
Andrea Leone$^{{\href{mailto:andrea.leone@ntnu.no}{\textrm{\Letter}}}1}$}
\centerline{\scshape
Per Thomas Moe$^{{\href{mailto:perthomas.moe@technipfmc.com}{\textrm{\Letter}}}2}$
Lasse Moldestad$^{{\href{mailto:lasse.moldestad@technipfmc.com}{\textrm{\Letter}}}2}$
}

\medskip

{\footnotesize
 \centerline{$^1$Norwegian University of Science and Technology, Norway}
} 

\medskip

{\footnotesize
 \centerline{$^2$TechnipFMC, Norway}
}

\smallskip



\begin{abstract}
Time series classification is of significant importance in monitoring structural systems. In this work, we investigate the use of supervised machine learning classification algorithms on simulated data based on a physical system with two states: Intact and Broken. We provide a comprehensive discussion of the preprocessing of temporal data, using measures of statistical dispersion and dimension reduction techniques. We present an intuitive baseline method and discuss its efficiency. We conclude with a comparison of the various methods based on different performance metrics, showing the advantage of using machine learning techniques as a tool in decision making.
\end{abstract}


\section{Introduction}\label{sec: intro}

In the offshore petroleum industry, drilling, completion and workover of subsea wells is usually performed by semi-submersible drilling rigs. A string of pipe sections extends from the rig to the subsea well and provides a conduit for fluid and tools. To prevent uncontrolled release of oil and gas to the environment this riser system includes a blowout preventer (BOP) directly on the top of the well. The BOP is a heavy steel structure with valves and allows for safe disconnect from the well if needed. A sketch of a BOP stack on a well can be seen in Figure \ref{fig:Stack with sensors} in Section \ref{sec: dataset}.

During operations wave forces acting on the rig, riser and BOP system induce cyclic loading in the uppermost part of the well (the wellhead). This will in turn cause fatigue damage and increase the risk of cracks to develop and grow in critical sections of the wellhead. A total or even partial loss of structural integrity and pressure control due to cracking of the wellhead must be prevented. For this reason great emphasis is placed on predicting and detecting changes in structural response.

During an operation sensor systems may continuously monitor riser and BOP accelerations and  the resulting bending moments applied to the wellhead. A systematic change in the relationship between these responses may be an indication of structural failure of the wellhead system. The change may, however, not be easily detectable for a human operator. This paper compares time series classification (TSC) methods for detecting changes in structural response. Several machine learning (ML) algorithms are trained on a synthetic, labelled, data set. Classification is performed either on the raw time series or by first making use of measures of variability of the data, like standard deviation (STD). Being able to classify a labelled data set with time series would serve as a proof of concept for training anomaly detection algorithms to detect cases where a crack occurs.

Our point of departure is a method relying on STD analysis of the data, which we will refer to as the baseline method. In this paper, we investigate and compare a range of alternative statistical approaches and ML techniques for binary classification of time series.  We use synthetic, but physically realistic data simulated by a state of the art commercial code and perform our analysis in a supervised learning setting.

The structure of this paper is as follows. In Section \ref{sec: dataset} we summarize the main characteristics of the data set and perform some preliminary analysis, which lays the basis for the following sections. We also introduce a formal definition of the supervised learning classification problem for the given time series data set. We conclude the section with a concise overview of Principal Component Analysis (PCA), one of the most popular dimension reduction techniques, whose theory goes back to Pearson \cite{pearson_1901} and Hotelling \cite{hotelling_1933}. We use \cite{jolliffe_2016} as our main reference. 

Sections \ref{sec:SotA}-\ref{sec: cnn} illustrate five methods to perform the classification task addressed in this work. For each method, we provide a brief description and report on the experimental results.

The baseline method is presented in Section \ref{sec:SotA}. This is mainly based on measures of variation of the values in the data set and on regression techniques. 

In Section \ref{sec:log_reg}, logistic regression (LogR) is used on the transformed data from Section \ref{sec: dataset}, combined with PCA. LogR was first introduced by Berkson \cite{berkson1944} in 1944 and applied to bioassay. Through the years it has been widely used in areas such as biology, medicine, psychology, finance and economics. It has become one of the most used classification algorithms, thanks to its simplicity, efficiency and interpretability, see e.g.  \cites{harrell2001regression, hosmer2013applied, menard2010logistic}. 

Section \ref{sec: dec_trees} covers Decision Trees (DTs), a popular supervised classification and regression technique introduced in the 1960s by Morgan and Sonquist in \cite{morgan1963problems}. New concepts, reviews of decision trees and their applications in different fields such as medicine, finance, environmental sciences, are presented in \cites{dt_medicine, dt_applications, dt_epidem}.  

Section \ref{sec: svm} illustrates how to use a Support Vector Machine (SVM) \cite{Boser92}, an ML algorithm for binary classification of data that continues to be widely popular due to its high performance and robustness to noise. Since the introduction of SVM in 1992 at AT\&T Bell Laboratories, it has been applied in fields such as medicine, biology, finance and technology \cite{Cervantes2020}.

The last method considered in this paper, investigated in Section \ref{sec: cnn}, belongs to the class of deep learning algorithms and uses a Convolutional Neural Network (CNN). Although CNNs were specifically introduced to work with image data \cite{lecun2015}, thus with input in the form of matrices (tabular data sets), they reached state of the art results also in other fields. In particular, they proved to be effective at capturing patterns in time series, making them among the most successful deep learning architectures for time series processing \cites{ismail2019, daesoo2023, sindre2023}.

In Section \ref{sec:comparison} we compare the methods based on different accuracy metrics and finally we provide conclusions and discuss research directions in Section \ref{sec:conclusion}.

\begin{table}[htbp]
\begin{center}
\setlength{\tabcolsep}{15pt}
\renewcommand{\arraystretch}{1.1}
\begin{tabular}{ll}
    \hline
    \multicolumn{2}{ c }{Nomenclature}\\
    \hline 
    \text{accx , accy} & \text{$x$ and $y$ component of the acceleration}\\
    \text{ASM} & \text{Attribute Selection Measure}\\
    \text{bmx, bmy} & \text{$x$ and $y$ component of the bending moment}\\
    \text{BOP} & \text{Blowout Preventer}\\
    \text{CNN} &  \text {Convolutional Neural Network } \\
    \text{DAS} & \text{Data Acquisition System}\\
    \text{DT}  & \text{Decision Tree}\\
    \text{DWS} & \text{Deep Water Strain }\\
    \text{FJ} & \text{Flex Joint}\\
    \text{LogR} & \text {Logistic Regression }  \\
    \text{ML} & \text {Machine Learning}  \\
    \text{MLP} & \text {Multi-layer Perceptron}  \\
    \text{PCA} & \text {Principal Component Analysis }  \\
    \text{SMU} & \text{Subsea Motion Units}\\
    \text{STD} & \text{Standard Deviation}\\
    \text{SVD} & \text{Singular Value Decomposition}\\
    \text{SVM} & \text{Support Vector Machine}\\
    \text{TSC} & \text{Time Series Classification}\\
    \text{WLR} & \text{Wire Load Relief}\\
    \hline
\end{tabular}
\end{center}
\caption{\label{tab:nomenclature}List of abbreviations and notations.}
\end{table}

\section{The data set under consideration}\label{sec: dataset}

The data set at hand is based on simulated data from the Orcaflex software package \cite{orcaflex}. This is done due to lack of measurements in the event of a well cracking. The simulated data is obtained from a three-dimensional finite element dynamic analysis in the time domain of the global riser, BOP and wellhead system. The system is exposed to realistic operational loads from a two-dimensional wave energy spectrum based on hindcast data gathered from representative operations. The two-dimensional sea state comprises 200 linear Airy wave components with different combinations of direction, frequency, and amplitude. In addition to waves, the system is exposed to a statistical median current profile for the same representative area. This is a unidirectional current with velocity varying with depth.

The riser, BOP and wellhead system is represented with one-dimensional line elements with six degrees of freedom. They are modelled with hydrodynamic, hydrostatic and structural properties aimed at giving realistic dynamic load exposure from the environment. This gives a realistic resulting dynamic load and deflection response.

The vessel used for the simulations is stationary, representing a bottom fixed operation vessel, and serves as a fixed reference for the top of the riser. The riser is in constant positive effective tension, with tension magnitude decreasing with water depth. The wellhead is modelled as a composition of line elements, and non-linear force displacement connections with nonlinear lateral force-displacement soil support in the form of P-Y curves, as is recommended practice, see \cite{dnv} and references therein.

In order to accurately capture the behaviour of intact and broken conditions, the model used in this study is adjusted to match the full three-dimensional solid finite element models of the broken and intact wellhead systems in soil, exposed to representative static loads. The simulation models for the global system and the wellhead calibration model are based on DNV-RP-E104, edition 2019-09 \cite{dnv}. 

\begin{figure}[htbp]
    \centering
    \resizebox{.8\textwidth}{!}{
        \input{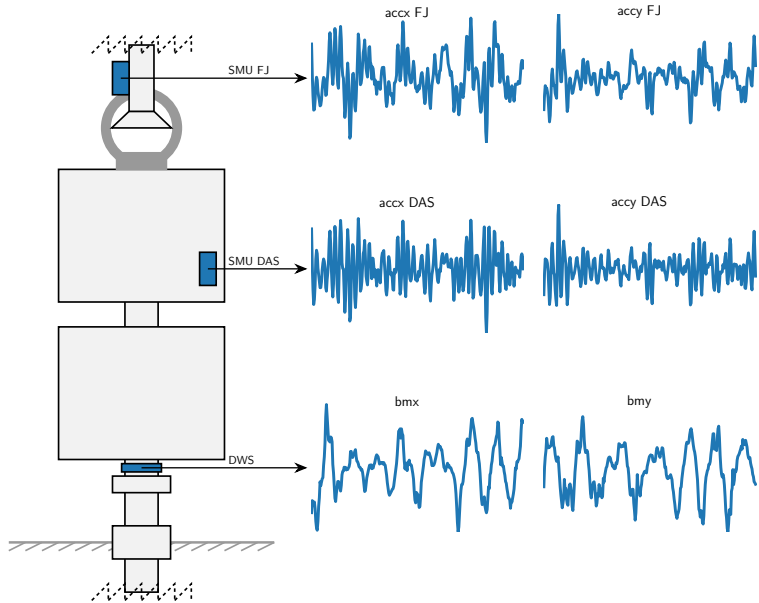}
        }
    \caption{Stack with sensors and corresponding data} 
    \label{fig:Stack with sensors}
\end{figure}

Sensors logging at $5$ Hz are simulated at likely sensor spots, see Figure \ref{fig:Stack with sensors}. For each sea state two one-hour data sets are created, each based on a simulation with and without a crack in the well, hereby referred to as \textit{broken} and \textit{intact}. The event where a crack occurs has to the authors' knowledge not been measured, nor is it simulated in the data set. Noise is added to the signal based on the sensor accuracy found in the data sheets relative to the in-operation sensors, with only \cite{SMU} being publicly available. Two other datasets are created with noise multiplied by 10 and 50, to test the robustness of the different methods. Hereby we refer to the three data sets as \emph{Noise 1, Noise 10,} and \emph{Noise 50}.

All of the data is normalised before applying any ML algorithms. Further details on data preprocessing can be found in Appendix \ref{sec:appendix_data_set}. Although the data observed in real-life operations may have more complex behaviour, we consider the artificial sensor data to suffice as a proof of concept that could be developed further in a later project with data gathered from the field.  

Before moving forward, we provide a formal definition of the supervised learning problem addressed in this work. We denote a univariate time series as $X_{{\text{uts}}}=\left[x_1, x_2, \ldots, x_n\right]$, which is an ordered set of real values $x_t$ indexed by integers $t = 1, 2, \dots, n$, with $x_t$ the value at the $t$-th discrete time point. We consider $X_{\text{uts}}$ as a column vector in $\mathbb{R}^n$. The simulations in our data set are associated with one-hour long measurements from 3 sensors, sampled at a rate of $5 \mathrm{~Hz}$. Each sensor outputs a signal for the $x$- and $y$-direction, hence we have a total of $m=6$ univariate time series with $n=18001$ data points. We can collect them in a multivariate time series, which we represent as a matrix
\begin{equation}\label{eq:x_multivariate}
X_{\text{mts}}=\left[X_{\text{uts}}^1, X_{\text{uts}}^2, \ldots, X_{\text{uts}}^6\right] \in \mathbb{R}^{n \times m}.
\end{equation}
We adopt a supervised testlearning approach to address the classification problem, as we have access to labelled data. More specifically, the dataset includes $N$ pairs $\mathcal{D}=\left\{\left(X_i, Y_i\right)\right\}_{i=1}^N$, where $X_i \in \mathcal{X}$ are input time series and $Y_i \in \mathcal{Y}$ the corresponding output variables. Here, $\mathcal{X}$ and $\mathcal{Y}$ denote the feature and label domains, respectively. Our aim is to approximate the mapping function 
\begin{equation} \label{eq:Fsp}
F: \mathcal{X} \rightarrow \mathcal{Y}, \quad Y_i=F\left(X_i\right),
\end{equation}
with sufficient accuracy so that we can make predictions about the output for any unseen input data. To this end, the data set is split $80\%-20\%$  into a training- and test-data set. A training procedure is performed on the former set by defining a loss function, that measures the distance between the predictions of the approximation to $F$ and the true labels, and a fitting optimisation algorithm. The accuracy of the approximation is then evaluated on the test set.

In this paper, we deal with a binary classification problem. We map input data into two discrete categories, intact and broken, to which we associate the labels $0$ and $1$ respectively, hence $\mathcal{Y} \equiv\{0,1\}$. Our \rez{original} data set consists of $N=103$ multivariate time series, 54 related to the intact case, and 49 to the broken one. Each of them is a collection of $18001$ values relative to $6$ signals, thus $\mathcal{X}\subset \mathbb{R}^{18001\times 6}$. The $6$ columns of each input data are called channels, and we will also refer to them as the number of input feature maps with a slightly abuse of terminology.

\subsection{Exploratory data analysis}

\begin{figure}[htbp]
     \centering
        \centering
        \includegraphics[width = \linewidth]{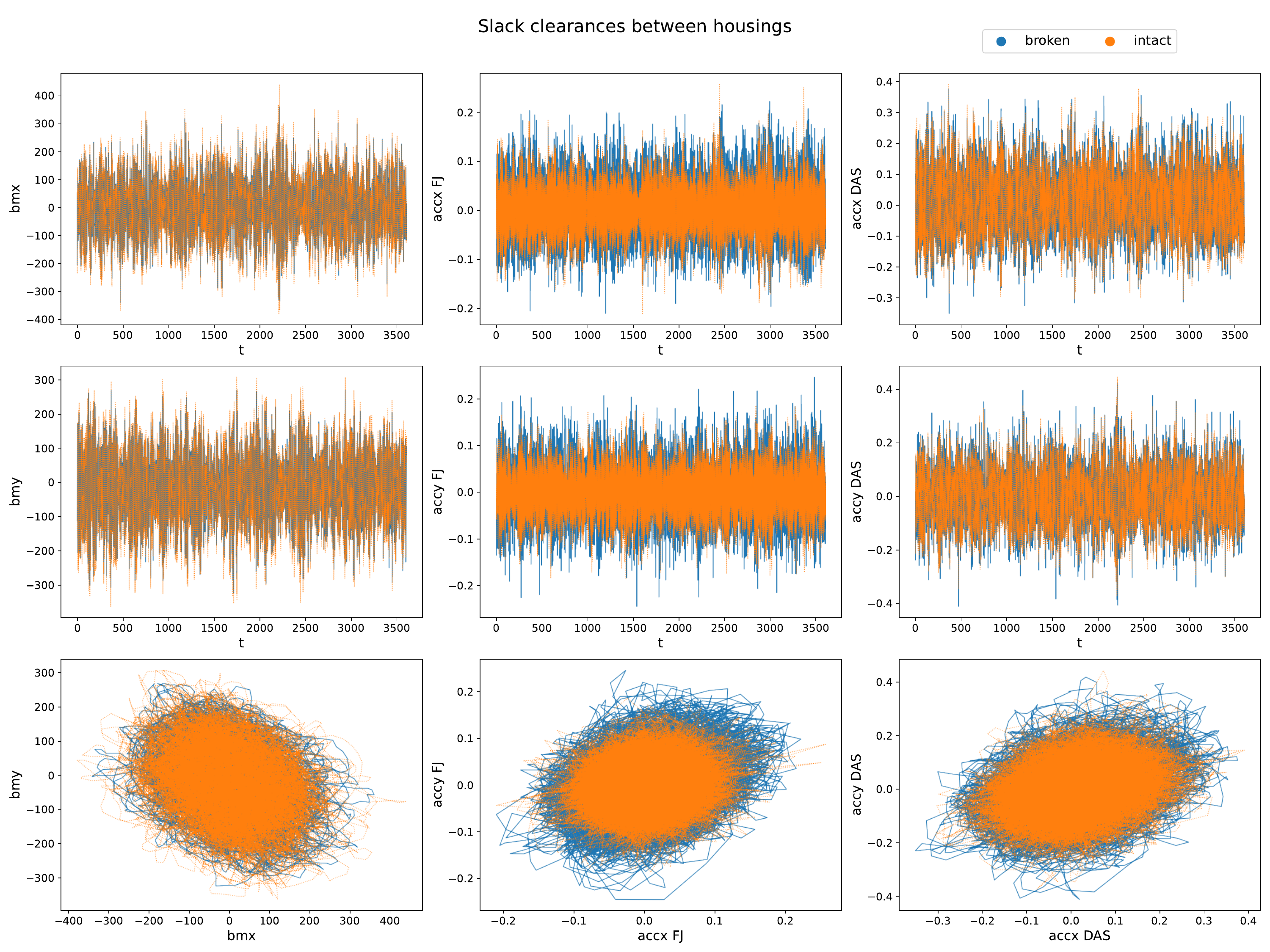}   
    \caption{Two 1-hour simulations from the dataset comparing a broken and intact well under similar conditions. Plots are given for the $x$ and $y$ component of the different physical measurements. The two top rows give the time series while the bottom row shows phase plots.}\label{fig:tightwholeseries}
\end{figure}

As we can see in Figure \ref{fig:tightwholeseries}, it is difficult to separate between an intact or broken well based on a single observation. We do however notice a difference in the spread of the data. This suggests to use a measure of dispersion when classifying.

\subsubsection{Standard deviations transformation}
To ensure that a crack in the wellhead is quickly noticed we look into classifying subintervals of the full data set. The simplest dispersion-based classification method consists of taking the standard deviation for each subinterval. More precisely, for each channel $m$, the standard deviation is calculated over \rez{one}-minute intervals. Therefore, each \rez{one}-minute interval with $m$ channels is mapped to a single data point with $m$ dimensions. \rez{One-minute intervals allow for updates of the well status at a satisfying frequency while being long enough to give reliable results.}

\begin{figure}[htbp]
     \centering
     \begin{subfigure}{0.59\textwidth}
        \centering
      \includegraphics[width=\linewidth]{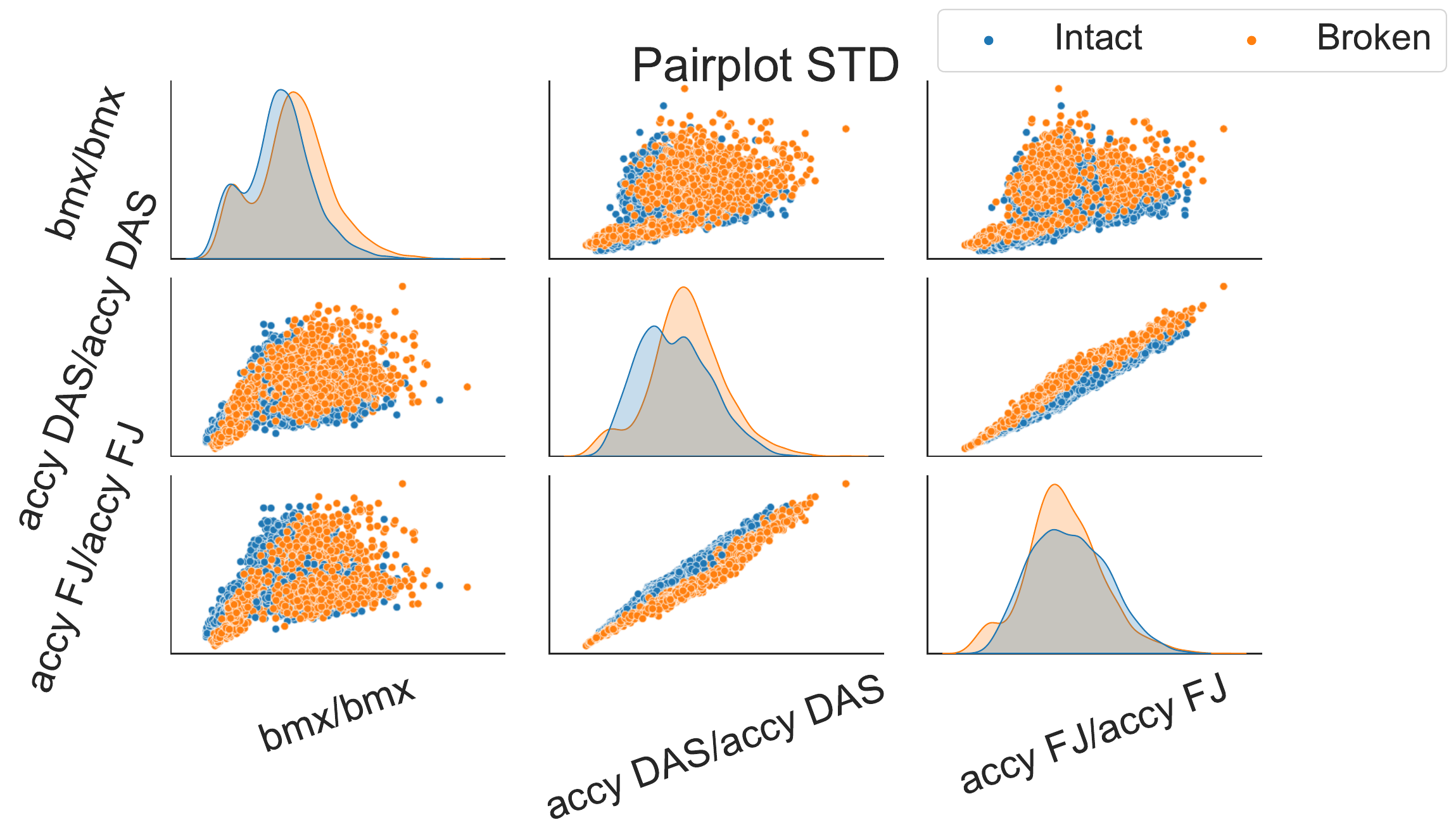}
     \end{subfigure}
     \hfill
     \begin{subfigure}{0.40\textwidth}
        \centering
      \includegraphics[width=\linewidth]{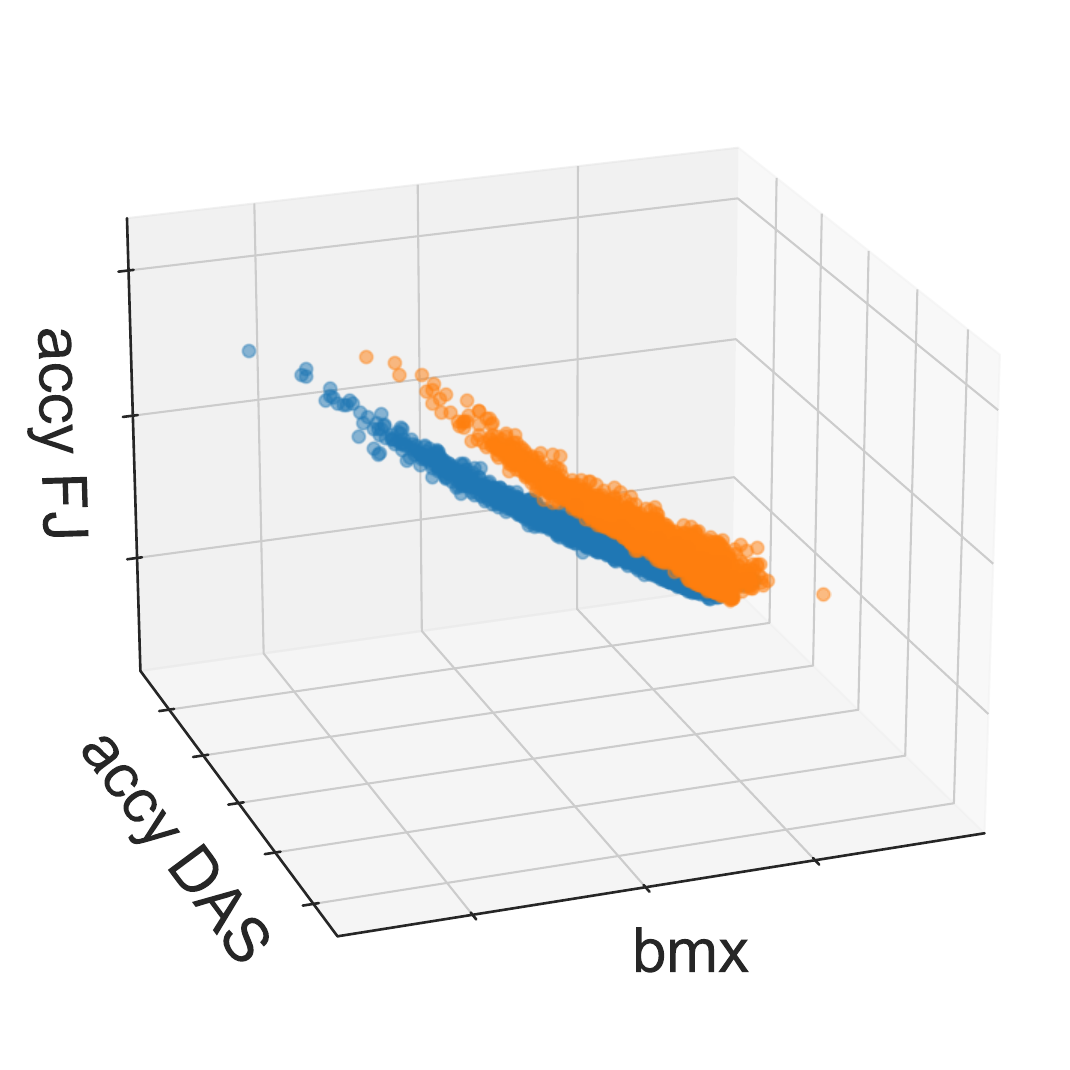}
     \end{subfigure}
     \hfill
    \caption{Pair plot showing of the scatter and distribution of data after a standard deviation transform (left). Plot visualizing the transformed data in 3 dimensions (right).}
    \label{fig:slack_std_pairplot}
\end{figure}

Applying this method to our data set gives us the point clouds found in Figure \ref{fig:slack_std_pairplot}. We immediately observe an increased ability to separate the two cases.

\subsubsection{Covariance transformation}

The standard deviation of the signals can be seen as a meaningful way of separating the data. This suggests that other statistical properties of the signals could be employed. Significant descriptive measures are provided by the covariance and correlation functions \cite{shumway2017time}, therefore we introduce the covariance matrix
 
\begin{equation}
\Sigma = \begin{bmatrix}
    \text{Var}(X_1) & \text{Cov}(X_1, X_2) & \cdots & \text{Cov}(X_1, X_n) \\
    \text{Cov}(X_2, X_1) & \text{Var}(X_2) & \cdots & \text{Cov}(X_2, X_n) \\
    \vdots  & \vdots  & \ddots & \vdots  \\
    \text{Cov}(X_n, X_1) & \text{Cov}(X_n, X_2) & \cdots & \text{Var}(X_n)
\end{bmatrix}.
\label{eq:covariance_matrix}
\end{equation}

Since we are working with standard deviations, we take the square root of the covariance matrix, given by
\begin{equation*}
    \Sigma^{\frac{1}{2}} = Q^\top \Lambda^{\frac{1}{2}} Q,
\end{equation*}
where $Q$ and $\Lambda$ store the eigenvectors and eigenvalues of $\Sigma$. As standard deviations are implicitly included in the covariance matrix, we highlight that the covariance transform expands the STD transform, thus adding more information.

It is worth noting that the covariance and correlation matrices are closely related since
\begin{equation}
    \text{Cor}(X) = \text{diag}(\Sigma)^{-\frac{1}{2}} ~\Sigma ~ \text{diag}(\Sigma)^{-\frac{1}{2}}.
\end{equation}
For most of the classification methods later presented, the covariance matrix is used, but in Section \ref{sec: cnn} correlation is indirectly utilized. 

Given the symmetry of the covariance matrix, only the upper triangular part of the matrix is used in the feature set. If $m$ defines the number of channels, one expects $\frac{1}{2}m (m+1)$ features. For the data set at hand this corresponds to $6$ or $21$ features, depending on whether one is using one or two physical directions from the sensor output.

\begin{figure}[htbp]
    \centering
    \includegraphics[width=\linewidth]{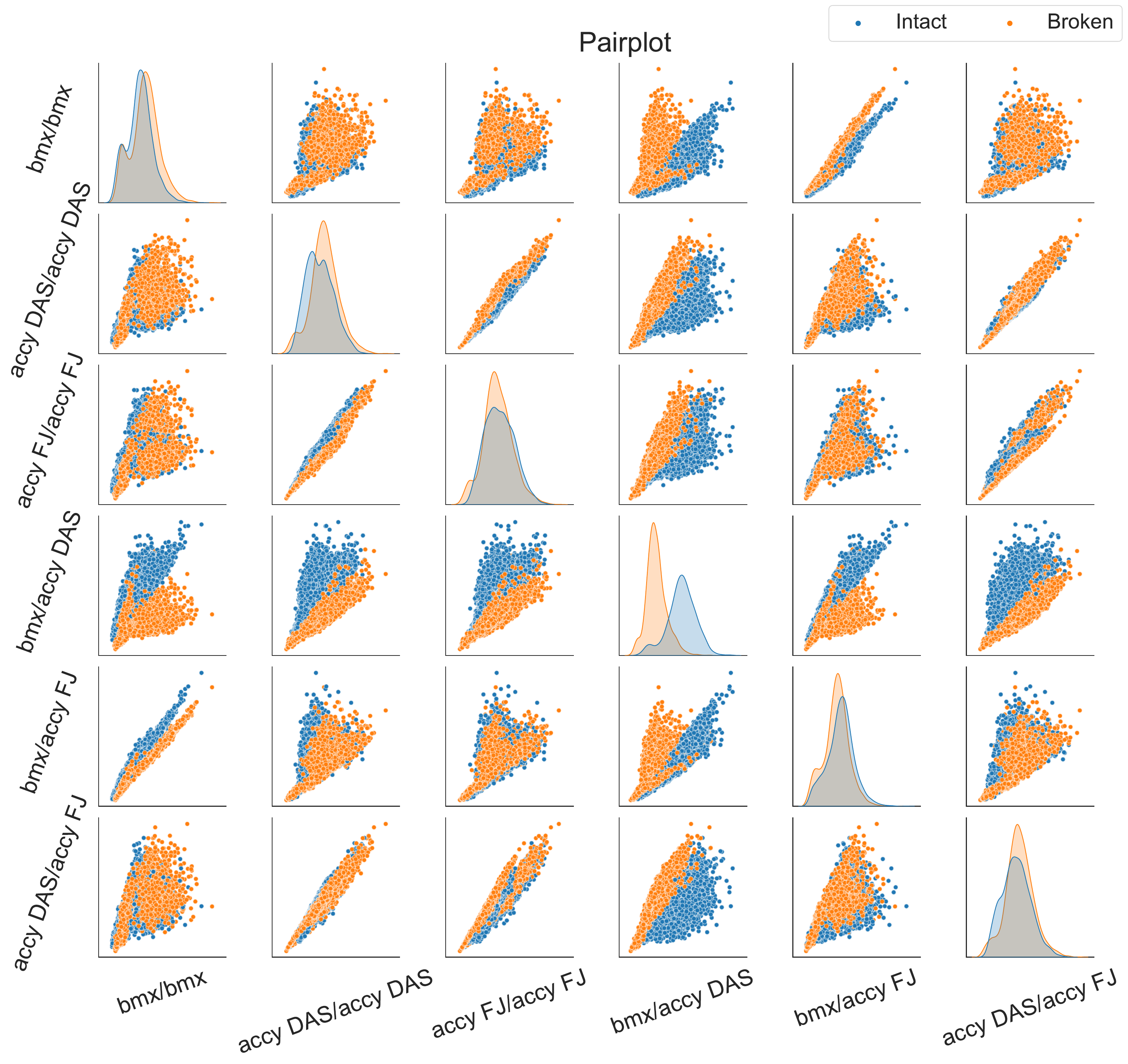}
    \caption{Pair plot of the data after using aforementioned covariance transform. For certain combinations the broken and intact cases separate quite well.}
    \label{fig:pair_plot_slack}
\end{figure}

In Figure \ref{fig:pair_plot_slack} we have restricted the data set to one physical direction and plotted a pairwise scatter plot to visualize the transformed data.
We observe an increased ability to distinguish between broken and intact compared to the standard deviation method, \rez{though the closeness of the point clouds still suggests difficulty in making correct classifications}. The main method of transforming the data will \rez{mainly} be through the use of the covariance matrix. 

The attentive reader can also observe that the top left $3\times 3$ block in Figure \ref{fig:pair_plot_slack} is similar to its corresponding figure with the standard deviation transform. This is to be expected, but underlines that the covariance matrix only adds relevant features.


\subsection{Principal Component Analysis}\label{subsec:pca_subsec}

PCA is an unsupervised dimension reduction technique that finds patterns or structures in the data and uses them to express the data in a compressed form.
This increases the interpretability of multidimensional data while preserving the maximum amount of information and enables its visualization. Preserving the maximum amount of information is equivalent to finding uncorrelated linear combinations of the original data set, called principal components, that successively maximize variance in addition to being uncorrelated with each other. Finding such new variables reduces to solving an eigenvalue-eigenvector problem. More precisely, a data set $\boldsymbol{X}$ is given as input to Algorithm \ref{alg:pca_alg}, provided below. \ergys{In this work, $\boldsymbol{X}$ will be either the STD- or the COV-transformed data.} The algorithm starts by solving an eigenvalue problem for the covariance matrix $\boldsymbol{\Sigma}$.
The $m \times m$ matrix $\boldsymbol{V}$ of eigenvectors diagonalizes the covariance matrix while $\boldsymbol{D}$ is the $m \times m$ diagonal matrix of eigenvalues of $\boldsymbol{\Sigma}$. 
The eigenvectors form a basis for the data and the eigenvalues represent the distribution of the information of the source data.

The goal is to choose a small enough subset of $d$ eigenvectors corresponding to the $d$ largest eigenvalues of $\boldsymbol{\Sigma}$. These will be the new basis vectors onto which we can project the data and still preserve a high quantity of information.
This is shown in the final step, where the $i$-th column of $\boldsymbol{P}$ is the projection of the data points onto the $i$-th principal component.

\begin{algorithm}[htbp]
    \begin{algorithmic}[1]
        \caption{Principal Component Analysis}
        \label{alg:pca_alg}
            \State \textbf{function} $\boldsymbol{P}$ = PCA($\boldsymbol{X}$): \Comment{$\boldsymbol{X}$ - input, $\boldsymbol{P}$ - output}
            \State $ \qquad \boldsymbol{X} \longrightarrow \frac{\boldsymbol{X} - \boldsymbol{\mu}}{\boldsymbol{\sigma}}$ \Comment{Normalize the data: $\boldsymbol{\mu}$ - mean, $\boldsymbol{\sigma}$ - standard deviation }
            \State $ \qquad \boldsymbol{\Sigma} = \frac{1}{n}\boldsymbol{X}^{\top}\boldsymbol{X}$ \Comment{Calculate the covariance matrix}
    
            \State $ \qquad \boldsymbol{V}^{T}\boldsymbol{\Sigma}\boldsymbol{V} = \boldsymbol{D}$ \Comment{Compute eigenvectors and eigenvalues of $\boldsymbol{\Sigma}$}

            \State $\qquad \boldsymbol{W} = \begin{bmatrix} \boldsymbol{w}_1, \boldsymbol{w}_2,\ldots, \ldots, \boldsymbol{w}_d \end{bmatrix}$ \Comment{Transformation matrix consisting on the {\begin{flushright}{first $d$ eigenvectors of $\boldsymbol{V}$ arranged in order of decreasing eigenvalues}\end{flushright}}}

            \State $\qquad \boldsymbol{P} = \boldsymbol{X}  \boldsymbol{W}$ \Comment{Project the data onto the new basis}

            \State \textbf{end function}
    \end{algorithmic}
\end{algorithm}

\begin{figure}[htbp]
    \centering
    \includegraphics[trim = 60 0 70 20, width = \textwidth]{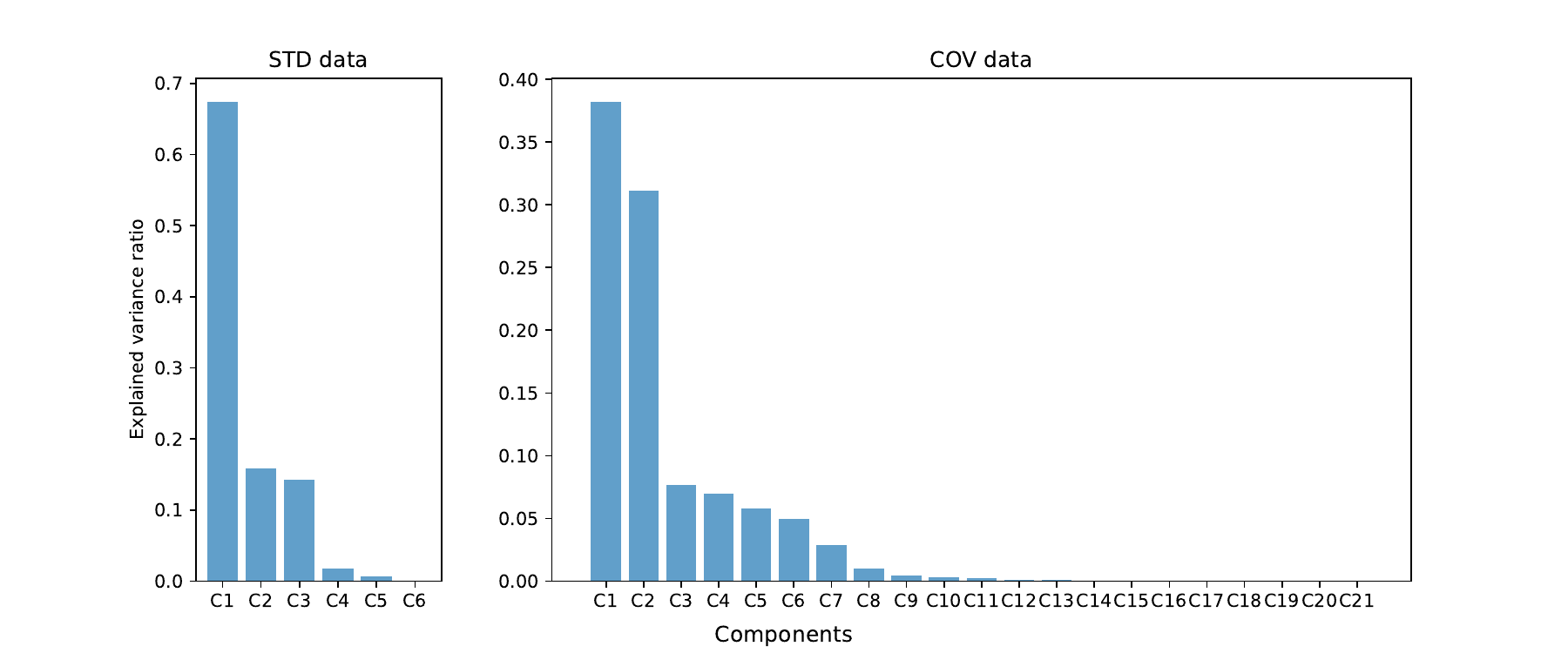}
    \caption{Ratio each component explains.}
    \label{fig:pcs_std_cov}
\end{figure}

Figure \ref{fig:pcs_std_cov} shows the ratio each component explains in the cases when the data is both STD- (left) and COV-transformed (right). In the first case, we see that most of the information is contained in the first 3 components, suggesting one only needs 3 PCs. In the second case, we see that the majority of information is contained in the first 7 components. The accuracy of the method increases along with the number of PCs.

\section{Baseline method} \label{sec:SotA}

The baseline method relies on standard deviation and regression, \rez{and is currently being used in production. It was designed to enable continuous human inspection and provide an intuitive visual representation of the current behaviour of the wellhead system. This is achieved by drawing regression lines on a monitor.} 

The method works by sliding a ten-minute window over each of the time series captured by the sensors. The window is split into one-minute intervals for which the standard deviation is calculated. Assume $m$ is the number of sensor channels and let $X \in R^{m\times 10}$ represent a matrix storing 10 calculated standard deviations for each channel. The method then relies on choosing two rows from $X$ and performing a linear regression. The two rows are typically chosen to be a bending moment and a flex joint acceleration corresponding to the same direction. The regression is given by the following equation
 
\begin{equation}
    \begin{bmatrix}
        \beta_0\\
        \beta_1
    \end{bmatrix}=
    \begin{bmatrix}
        x^\top x & x^\top \mathbb{1}\\
        \mathbb{1}^\top x & \mathbb{1}^\top \mathbb{1}
    \end{bmatrix}^{-1}
        \begin{bmatrix}
        x^\top\\
        \mathbb{1}^\top
    \end{bmatrix}
    y,
    \label{eq:linear regression}
\end{equation}
where $\beta_0$ is the intercept and $\beta_1$ is the incline of the regression line, respectively. 
The ten-minute time window is then moved one time step forward and a new line is drawn. The time step is user defined and is typically set to one minute. 

\begin{figure}[htbp]
    \centering
    \includegraphics[width=\linewidth]{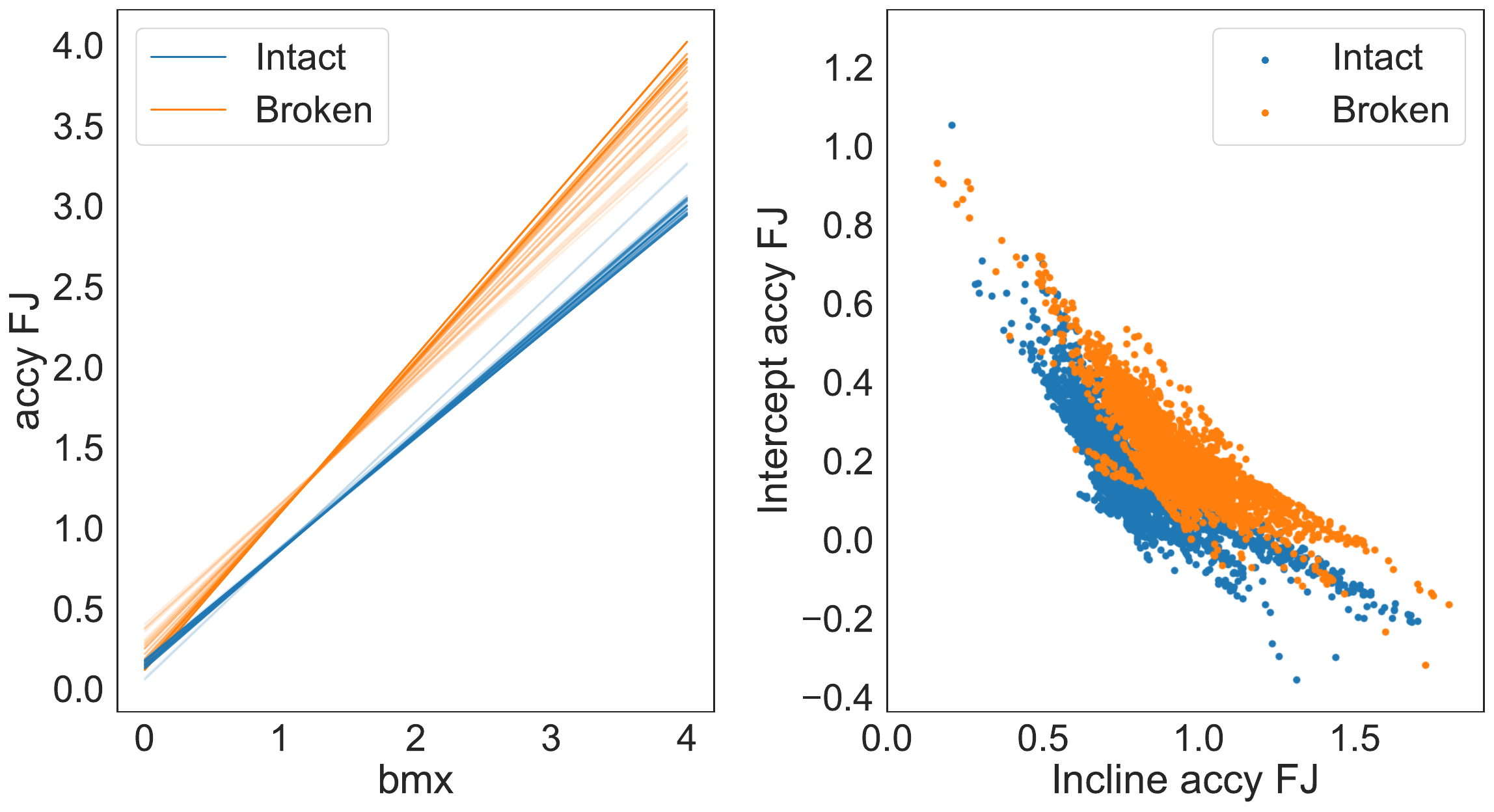}
    \caption{\rey{Left: Visualization of the lines capturing the relation between the standard deviation of accelerations in the flex-joint and wellhead bending moments using linear regression. The lines are meant to be displayed on a vessel's monitor and gradually fade over time highlighting the most recent behaviour.}  Right: Distribution of data for the baseline method.}
    \label{fig:SotA plot}
\end{figure}

Any significant change between the drawn lines indicates a change in behaviour of the system. Therefore, the occurrence of a crack should be detectable through continuous monitoring of the data. An example of the lines for the cases of a broken/intact well, simulated in a similar environment, can be seen in Figure \ref{fig:SotA plot}. The event where a crack occurs has to the authors' knowledge not been measured, nor is it simulated in the data set.

To analyse the method further, we look at the distribution of the intercept and incline of the regression line. The plot to the right in Figure \ref{fig:SotA plot} illustrates how data points are separated based on whether the well is broken or intact. One may observe a noticeable separation of data, but there is overlap and they lie very close. \rez{Similarly to what was observed in Figure \ref{fig:pair_plot_slack},} the closeness of the two distributions suggests difficulty in detecting change in behaviour implying \rez{difficulty in classifying the data}.

An important feature with this baseline method is the temporal dependence between the lines (left) or points (right) in Figure \ref{fig:SotA plot}. Given the lack of recorded cracking events, we can only speculate on its efficiency. We could however expect a crack to cause the data points to move from their positions in the point cloud representing intact cases to a similar position in the point cloud representing broken. 
However, given the constraints of our data set, we limit ourselves to examine individual data points whenever a method of dispersion is used.

As a final remark, the linear regression is related to the covariance transform. This becomes clearer when rewriting equation \eqref{eq:linear regression} using the mean, variance and covariance as follows
 
\begin{align*}
    b_0 &= \mu_y - \frac{\mu_x \Cov(x, y)}{\text{Var}(x)} ,\\
    b_1 &= \frac{\Cov(x,y)}{\text{Var}(x)}.
\end{align*}
From the equation we read that the baseline method essentially approximates the point clouds from a subplot, depending on the sensor chosen, in Figure \ref{fig:pair_plot_slack} with a linear regression. The method does however suffer from high uncertainty due to the small set of samples in each prediction.

\section{Logistic Regression}\label{sec:log_reg}

Given the reduced feature matrix $\boldsymbol{P}$ from Algorithm \ref{alg:pca_alg} in Section \ref{subsec:pca_subsec}, binary LogR uses a regression technique to solve the two-class classification problem with the class variable \textit{Target} $= \{$\textit{Broken, Intact}$\}$ by modelling the class probability $P=\operatorname{Pr}(\textit{Target} = \textit{Intact} \mid \boldsymbol{P})$ as
\begin{equation}\label{class_prob_log}
\log \frac{P}{1-P}={\beta}_0+{\beta}^{\top}\boldsymbol{P},
\end{equation}
with an intercept ${\beta}_0$ and a parameter vector ${\beta}$. The class probability is defined as
\begin{equation}\label{class_prob_p}
P=\frac{\exp \left({\beta}_0+{\beta}^{\top} \boldsymbol{P}\right)}{1+\exp \left({\beta}_0+{\beta}^{\top} \boldsymbol{P}\right)}.
\end{equation}
Fitting a logistic regression model means estimating \ergys{the intercept ${\beta}_0$ and } the parameter vector ${\beta}$. In our experiments, this is done via the \texttt{LogisticRegression} from \texttt{sklearn.linear\_model} \ergys{with all parameters set to their defaults}.

\subsection{Experiments}\label{subsec:experiments_pca_logr}
\begin{figure}[htbp]
    \centering
    \includegraphics[width = \textwidth]{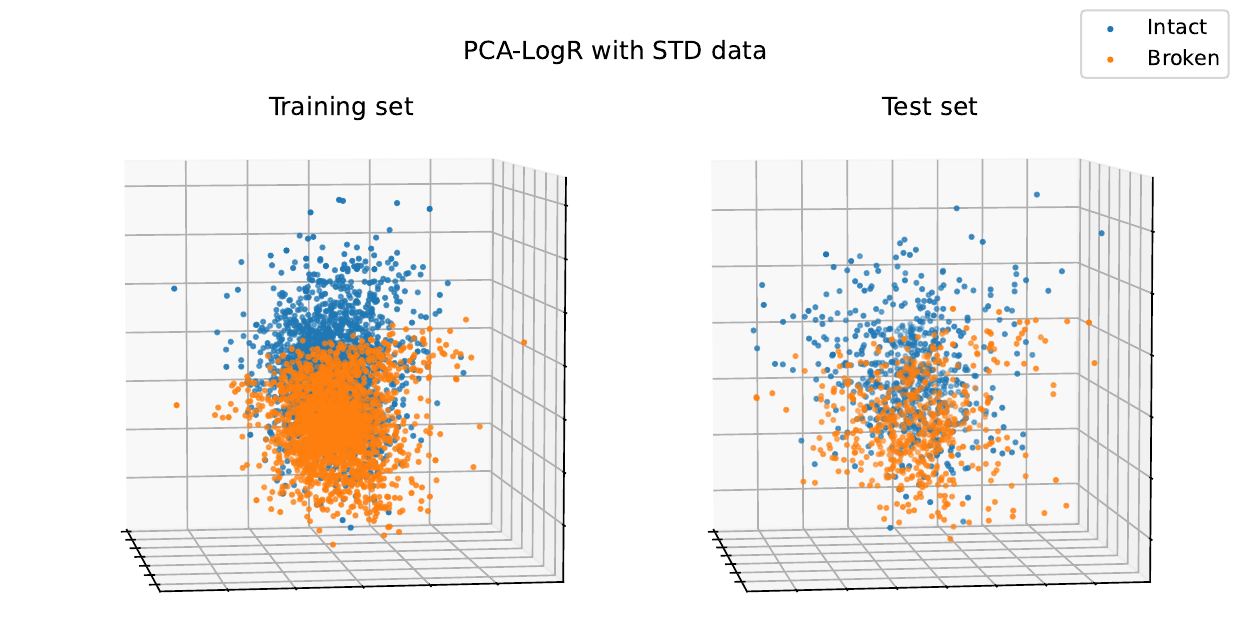}
    \caption{Classification of the STD data \ergys{from the \emph{Noise 1} data set} with 3 principal components.}
    \label{fig:train_test_set_pca_std}
\end{figure}

\begin{figure}[htbp]
    \centering
    \includegraphics[width = \textwidth]{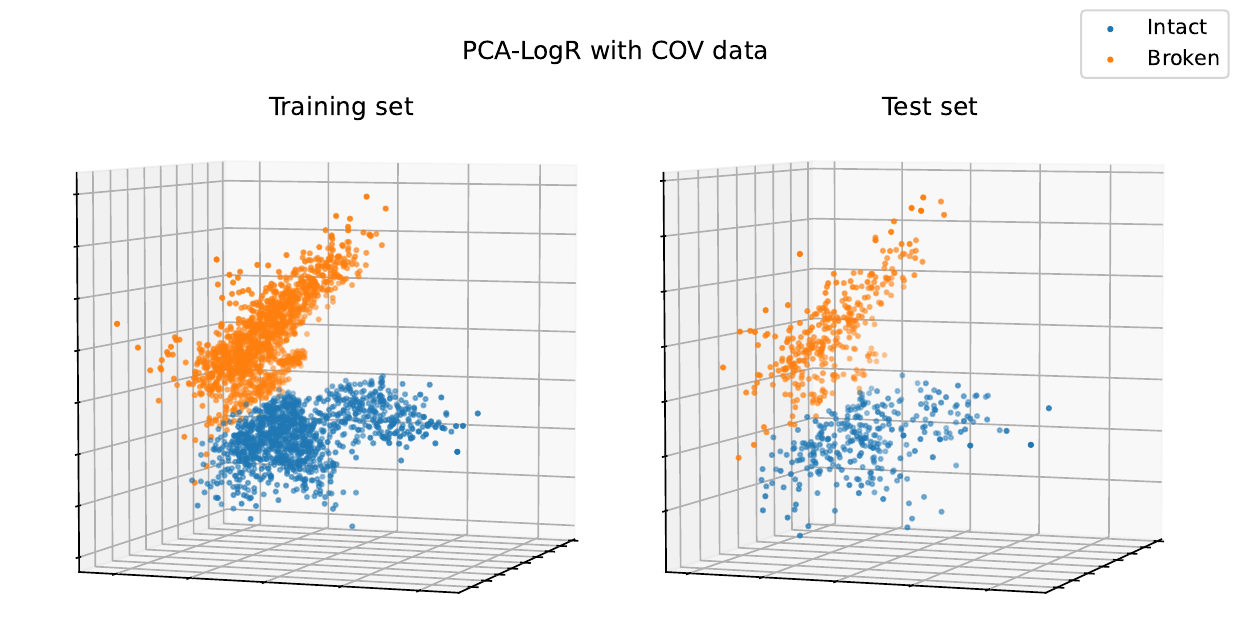}
    \caption{Classification of the COV data \ergys{from the \emph{Noise 1} data set} with 4 principal components.  The 3D visualization is made with 3 components.}
    \label{fig:train_test_set_pca_cov}
\end{figure}

In this subsection, we show experiments performed by applying LogR to the reduced feature data set, the output of Algorithm \ref{alg:pca_alg}. \rey{We utilize the existing implementation of PCA outlined in Algorithm \ref{alg:pca_alg}, available through the function \texttt{PCA} from \texttt{sklearn.decomposition}.}
We fit \texttt{LogisticRegression} to the training set and use the \texttt{predict} function to predict the test set result. The LogR-PCA approach is applied to both the STD- and the COV-transformed data \ergys{from the data set \emph{Noise 1}, \emph{Noise 10}, and \emph{Noise 50}, respectively}. 
\ergys{For the STD-transformed data, we test the accuracy of the method with the number of PCs going from 1 to 6. In the case of the COV-transformed data, we test for PCs from 1 to 7, since we see from Figure \ref{fig:pcs_std_cov} that those contain the majority of information. The accuracy of the method in such scenarios, measured with \texttt{accuracy.score} of \texttt{sklearn.metrics} as the ratio of correctly predicted samples to the total number of samples, is reported in Table \ref{tab:pcs_cov_vs_accuracy}. We see that for the same number of PCs, a higher level of noise leads to a lower accuracy. Hence, to achieve high accuracy even with noisy data, it is necessary to increase the number of PCs. In Figures \ref{fig:train_test_set_pca_std} and \ref{fig:train_test_set_pca_cov}, the classification of the time series in the training and test sets is shown for both the STD- and the COV-transformed \emph{Noise 1} data.}

\begin{table}[htbp]
\begin{center}
\setlength{\tabcolsep}{6pt}
\renewcommand{\arraystretch}{1.1}
\begin{tabular}{lllllllll}
    \hline 
    \multicolumn{2}{c}{\multirow{3}{*}{$\begin{array}{c}
    \text {Data set and} \\
    \text {data transformation}
    \end{array}$}} & \multicolumn{7}{c}{Accuracy (\%)} \\ \cline{3-9}
    & & \multicolumn{7}{c}{Number of PCs} \\
    \cline{3-9}
    & & 1 & 2 & 3 & 4 & 5 & 6 & 7
    \\
    \hline
    \multirow{2}{*}{Noise 1} & STD & 
    55.99 & 54.53 & 69.26 & 69.17 & 98.46 & 98.62 & -\\
    \cline{2-9} 
    & COV  & 55.24 & 55.56 & 65.88 & \textbf{99.69} & 99.84 & 100 & 100\\   
    \hline
    \multirow{2}{*}{Noise 10} & STD & 
    55.66 & 54.53 & 69.17 & 69.17 & 98.14 & 98.14 & - \\
    \cline{2-9} 
    & COV  & 55.56 & 55.87 & 64.16 & \textbf{99.53} & 99.84 & 99.84 & 99.84 \\   
    \hline
    \multirow{2}{*}{Noise 50} & STD & 
    54.29  & 54.21 & 68.77 & 69.01 & 89.97 & 91.26 & - \\
    \cline{2-9} 
    & COV  & 55.56 & 56.81 & 54.93 & \textbf{79.34} & 91.06 & 95.62 & 96.09 \\   
    \hline
\end{tabular}
\end{center}
\caption{\label{tab:pcs_cov_vs_accuracy}Accuracy of LogR-PCA applied to the STD and COV data from the different data sets with different number of PCs. In bold are marked the scenarios that will be reported in Table \ref{tab:comp_table} for comparison purposes.}
\end{table}

\section{Decision trees}\label{sec: dec_trees}

\begin{figure}[htbp]
    \centering
    \includegraphics[width = \textwidth]{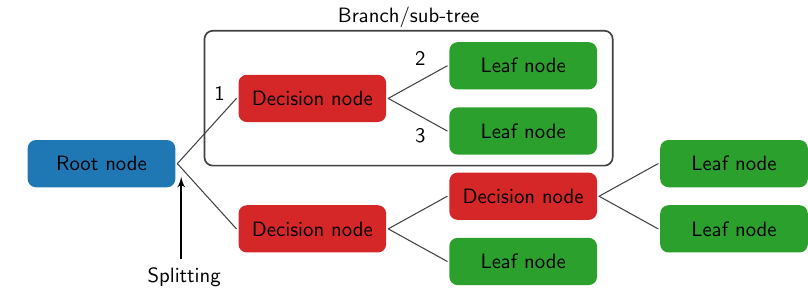}
    \caption{Example of a horizontal decision tree with depth 3. Node 1 is the parent node of nodes 2 and 3.}
    \label{fig:dt_scheme}
\end{figure}

A decision tree (DT) is a model that predicts the value of a target variable by learning simple decision rules inferred from the data features. Given a labelled data set, the model categorizes the data into purer subsets, \rez{ i.e., subsets consisting of highly homogeneous data,} based on a set of if-else conditions. 
One can think of a DT as a piece-wise constant approximation of the final classification. Figure \ref{fig:dt_scheme} provides some common terminology and illustrates the idea behind decision trees.

The quality of the splitting, \rez{which refers} to the purity of the resulting nodes, is measured with Attribute Selection Measure (ASM) techniques. The root node feature is selected based on the results of the ASM, and the procedure is repeated until a node cannot be split into sub-nodes, i.e., until it becomes a leaf node. 
More specifically, starting from the root node, we evaluate how poorly each feature splits the data into the correct classes, intact or broken. The feature resulting in the lowest impurity is chosen as the best feature for splitting the current node. This is repeated for each subsequent node.
There exist two typical ASM techniques for measuring purity, namely \textit{Gini impurity} or \textit{Gini index} and \textit{information entropy} or \textit{information gain}, \cites{mola1997fast, rokach2005top, tangirala2020evaluating}.

The Gini impurity, or the Gini index, ($GI$) measures the probability of a particular variable being wrongly classified when randomly chosen. In node $d$, \rey{the quantity} $GI$ is calculated as 
\begin{equation}\label{eq: gini_index}
GI_d = 1 - \sum^l_{k=1} p^2_{d,k},
\end{equation}
where $p_{d,k}$ denotes the probability of an object in node $d$ being classified into the class $k = 1, \ldots, l$. 
When the parent node $d$ is split, based on a feature $f$, into $m$ nodes $d_i, i = 1, \ldots, m$, the resulting GI is calculated as the following weighted average:
\begin{equation}\label{eq: gini_index_f}
GI_d|_f = \sum_{i = 1}^m \frac{|d_i|}{|d|}GI_{d_i},
\end{equation}
where $|\cdot|$ denotes the number of data in a node and $GI_{d_i}$ are calculated as in Equation \eqref{eq: gini_index}.
When this criterion is used for the selection of the root node feature, the 
feature with the smallest $GI$ is selected. The lower the $GI$ of a node, the closer the node is to being a leaf node. The $GI$ of a pure node is 0.

The information Gain $(IG$) criterion is based on the entropy ($E$) measured  in each node, which decreases as the purity of the node increases. A pure node has entropy 0. In node $d$, \rey{the quantity} $E$ is calculated as: 
\begin{equation}\label{eq: entropy}
E_d = - \sum^l_{\substack{k=1\\ p_{d,k} \neq 0}} p_{d,k} \log_2(p_{d,k}), 
\end{equation}
where $p_{d,k}$ is as before. 
The information Gain $(IG$) measures the decrease in entropy by computing the difference between entropy before the split and average entropy after the split of the node, based on the chosen feature. 
Suppose, similarly to above, that the parent node $d$ is split, based on a feature $f$, into $m$ nodes $d_i, i = 1, \ldots, m$. Then $IG$ of the feature $f$ in node $d$ is calculated as:
\begin{equation}\label{eq: inf_gain}
IG_d|_f = E_d - \sum_{i = 1}^m \frac{|d_i|}{|d|}E_{d_i},
\end{equation}
where $E_{d_i}$ are calculated as in Equation \eqref{eq: entropy}.
The feature yielding the highest $IG$ is chosen as the splitting feature for the node in consideration.

\begin{figure}[htbp]
    \centering
    \includegraphics[width = \textwidth]{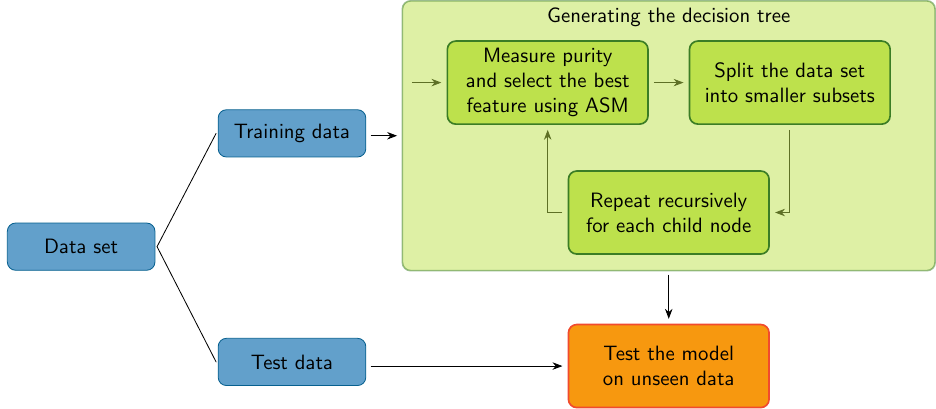}
    \caption{Decision tree algorithm \ergys{illustrated as in \cite{Kim2022}}.}
    \label{fig:dec_tree_alg}
\end{figure}

There is no big difference between Gini impurity and entropy when it comes to efficiency, see \cite{raileanu2004theoretical}. The choice varies significantly on the particular circumstances and the data set. One advantage of the GI to the entropy approach is that it does not involve logarithms, which are expensive from a computational point of view. Figure \ref{fig:dec_tree_alg} shows how the DT algorithm works. 

One common difficulty for DTs is overfitting. It can be prevented in two common ways, namely constraining the tree size and pruning the tree, often known as pre-pruning and post-pruning, respectively. Pre-pruning is done by controlling the following parameters: the minimum number of samples required for a node to split, the minimum number of samples for a leaf node, the maximum number of leaf nodes, the maximum depth of the tree, the maximum number of features to consider while searching for the best split. In post-pruning, nodes and subtrees are replaced with leaves to reduce the complexity of the tree.

\subsection{Experiments}\label{subsec: experiments_dts}
\begin{figure}[htbp]
\centering
    \includegraphics[width = \linewidth]{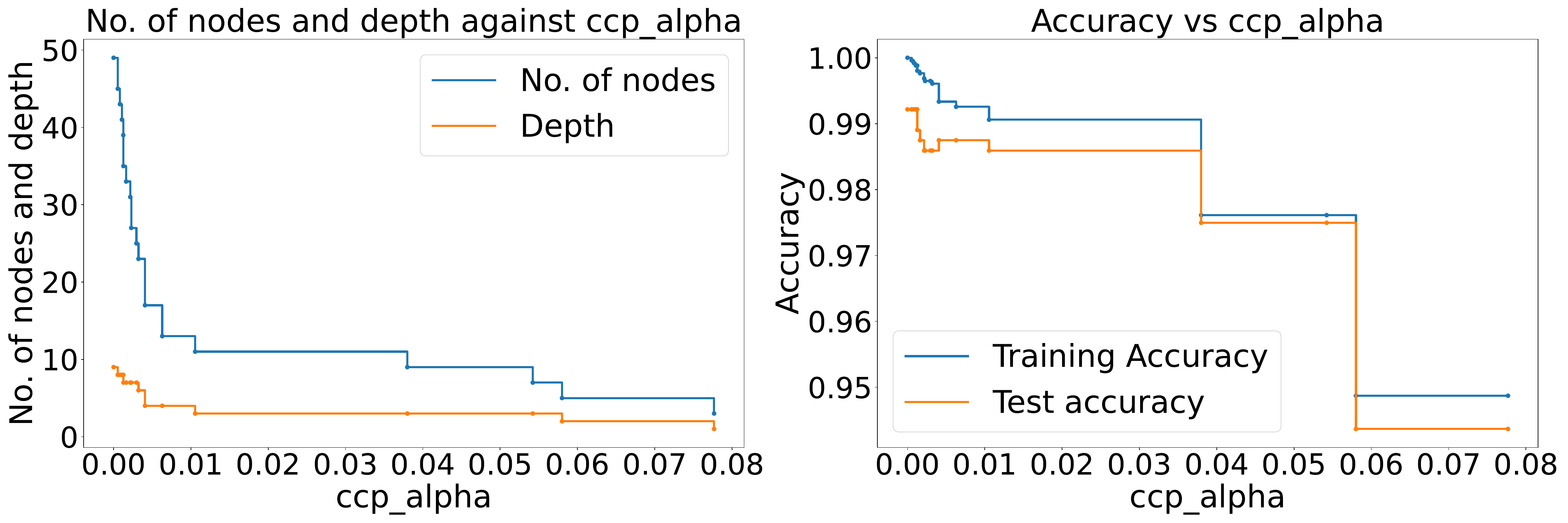}
    \caption{The effect of \texttt{ccp\_alpha} on the structure and the accuracy of the tree. Scenario: Noise 1, COV-PCA(4), Entropy (marked in bold in Table \ref{tab:dt_table}.)}
    \label{fig:choosing_alpha_dt_cov_pca_4}
\end{figure}

In the numerical experiments, the trees are generated using the function \texttt{tree.DecisionClassifier} from \texttt{sklearn} of \texttt{Python}, where one can choose between \texttt{entropy} or \texttt{Gini} splitting \texttt{criterion}, and they are displayed using the visualization tool of the \texttt{tree} class. \ergys{\texttt{sklearn} uses an optimised version of the CART algorithm \cite{cart_algorithm} which uses \texttt{gini} as splitting criterion and considers a binary split for each attribute. When \texttt{entropy} is chosen as splitting criterion, the ID3 algorithm \cite{id3_algorithm} is used.} 
Pre-pruning is performed using the function \texttt{GridSearchCV} from \texttt{sklearn}, which does a thorough search for an estimator over the specific set of parameter values described in the previous section. For the post-pruning, the \texttt{cost\_complexity\_pruning\_path} function is used, which is parameterized by the cost complexity parameter \texttt{ccp\_alpha}. By increasing the value of \texttt{ccp\_alpha}, the number of pruned nodes increases, and consequently the accuracy decreases, see Figure \ref{fig:choosing_alpha_dt_cov_pca_4}. Therefore, one has to make a clever choice of this parameter in order to have significant results. 
One has to accept a decrease in accuracy in return for a significant reduction in tree complexity.

\begin{figure}[htbp]
\centering
    \includegraphics[width = 0.5\linewidth]{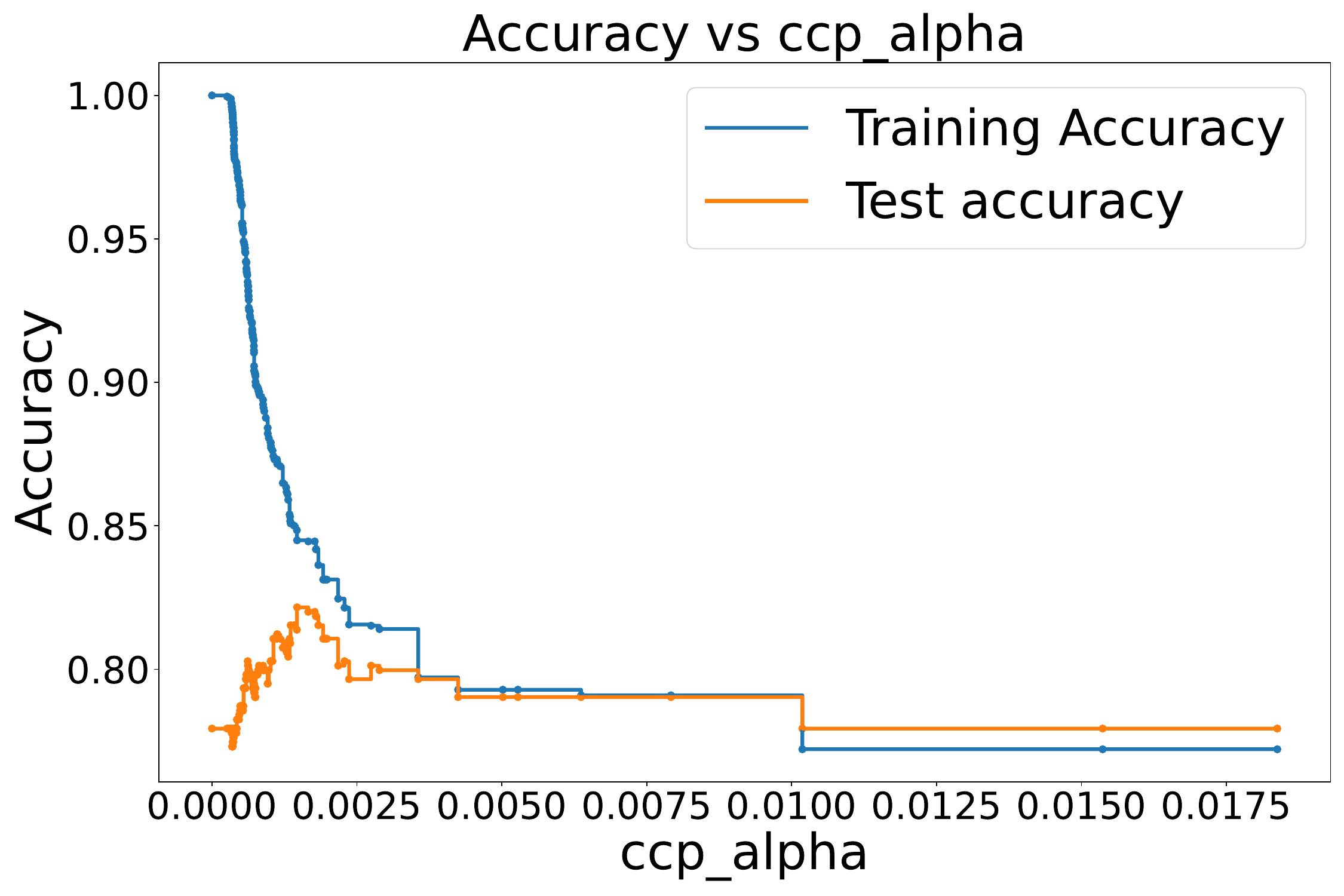}
    \caption{The effect of post-pruning in the reduction of overfitting. Scenario: Noise 50, COV-PCA(4), Gini (bottom-right block of Table \ref{tab:dt_table}.)}
    \label{fig:post_pruning_overfitting}
\end{figure}

\begin{figure}[htbp]
    \centering

    \includegraphics[width = \textwidth]{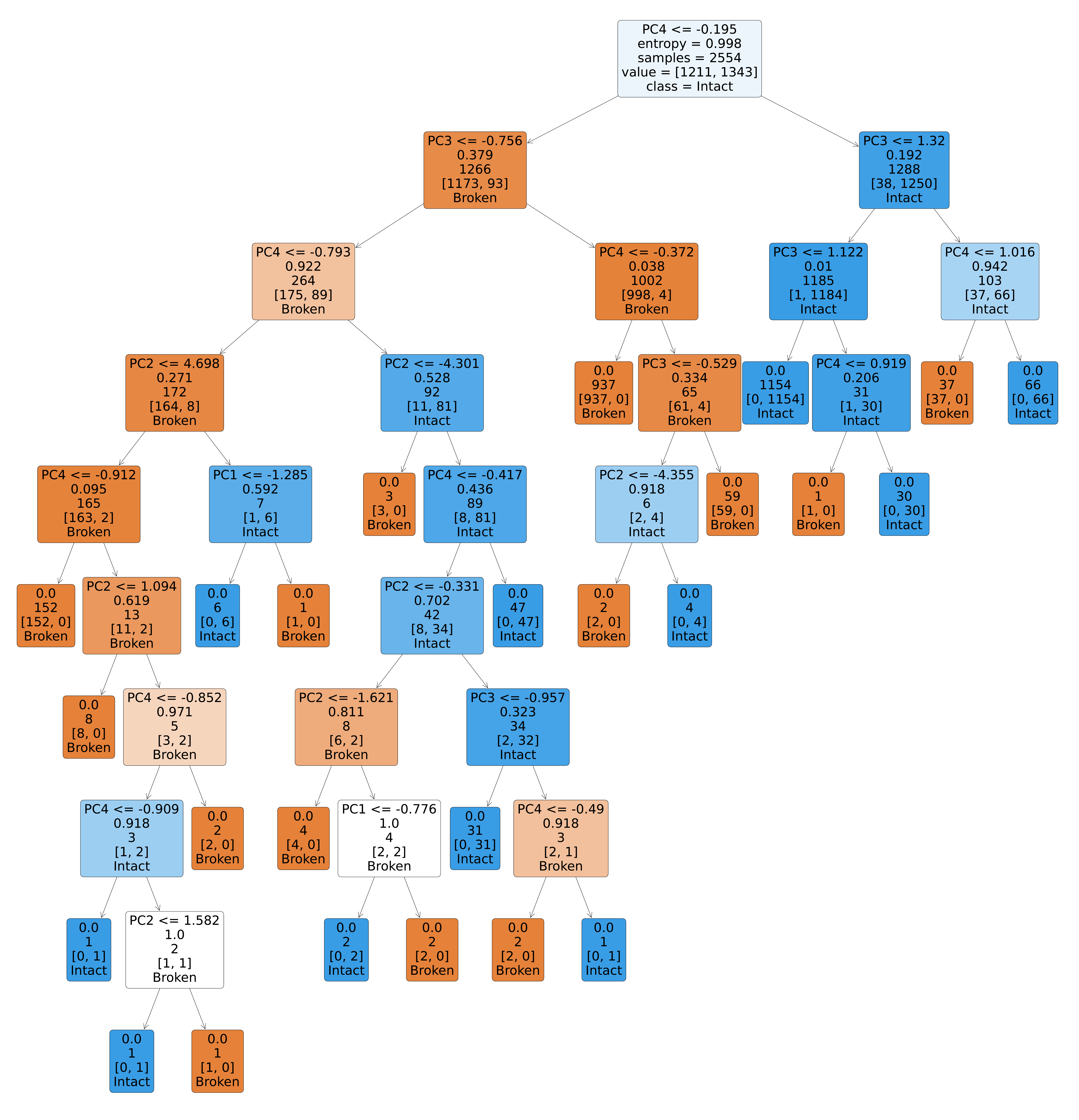}
    \caption{DT generated with entropy as splitting criterion on the data set consisting of the first four PCs of the COV data. \rey{Blue and orange are used for intact and broken, respectively. A light colour indicates a high entropy, an intense colour a low entropy.
    }}
    \label{fig:overlay_dts_horizontal}
\end{figure}

\begin{figure}[htbp]
    \centering
    \includegraphics[width = 0.8\textwidth]{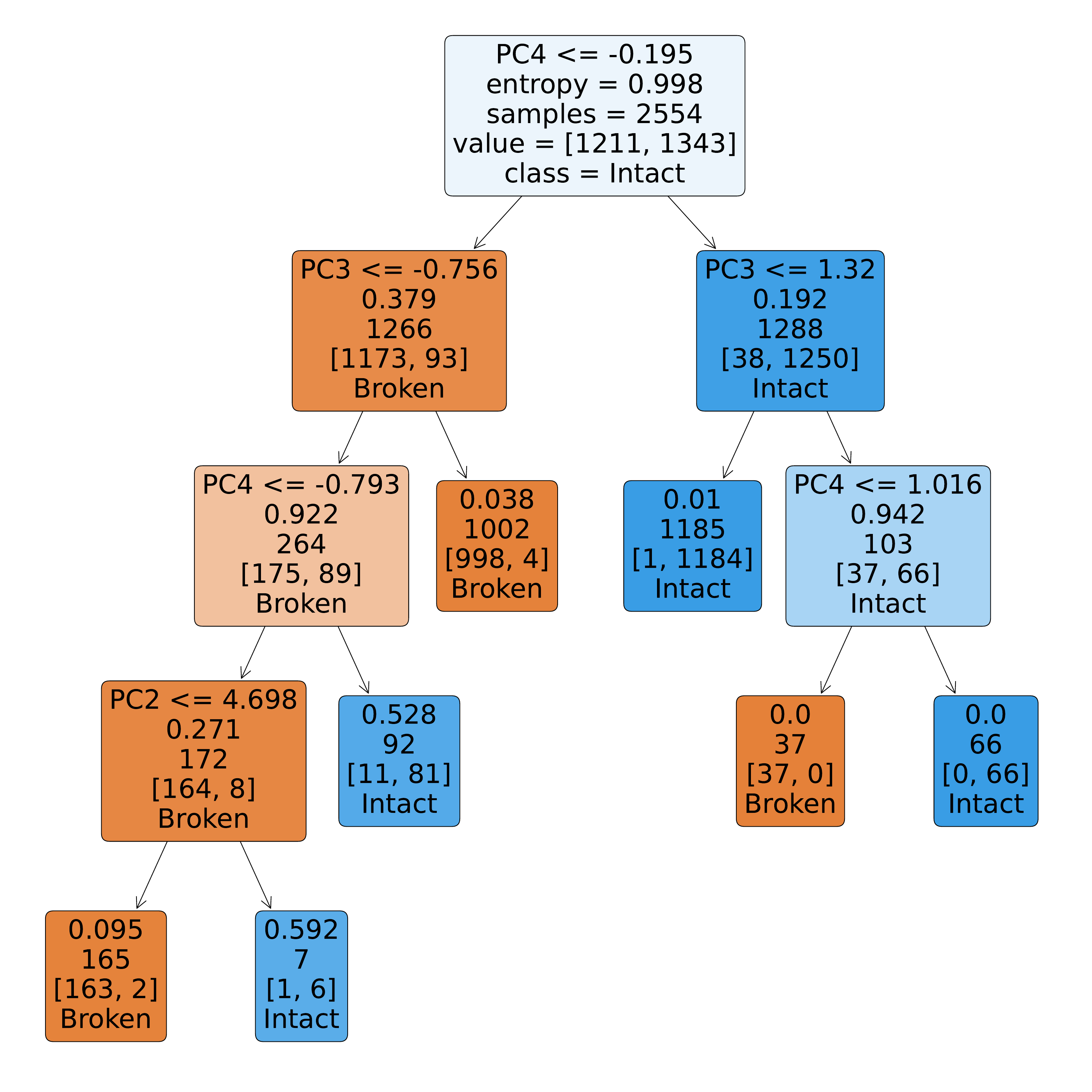}
    \caption{The same DT as in Figure \ref{fig:overlay_dts_horizontal} post pruned with \texttt{ccp\_alpha} = 0.01.}
    \label{fig:dt_prunned_rotated}
\end{figure}

A series of experiments are run on different scenarios and the results are reported in Table \ref{tab:dt_table}. \rey{The hyperparameter range for the pre-pruning and choice of the $\alpha$ for the post-pruning of the DTs, used to obtain the results reported in Table \ref{tab:dt_table}, is provided in Appendix \ref{sec:supp_mat_dts}.} \rex{There is no sign of overfitting of the model in the case of \emph{Noise 1} and \emph{Noise 10} but we notice overfitting in the case of \emph{Noise 50}. We can also see the positive effect of pruning in the reduction of overfitting, in particular when post-pruning. In Figure \ref{fig:post_pruning_overfitting}, this is shown for the Noise 50, COV-PCA(4) data split with Gini criterion, corresponding to the values in the bottom-right block in Table \ref{tab:dt_table}.} In Figure \ref{fig:overlay_dts_horizontal}, we show the tree generated with \texttt{entropy} as splitting criterion applied to the data set consisting of the first four PCs of the COV data. In Figure \ref{fig:dt_prunned_rotated}, the post-pruned version of the same tree with \texttt{ccp\_alpha} $= 0.01$ is shown. The value for \texttt{ccp\_alpha} is suitably chosen in Figure \ref{fig:choosing_alpha_dt_cov_pca_4}. For presentation purposes, the labels are shown only on the root node. The root and decision nodes include the following information: the feature in the data set that best divides the data, the value of the entropy, the number of the samples, their division into the classes and the dominant class, respectively. Leaf nodes are pure and there is no decision to be made.

\newcolumntype{g}{>{\columncolor{mygray}}l}
\begin{sidewaystable}[htbp]
\vspace*{13cm} 
\centering
\setlength{\tabcolsep}{1pt}
\renewcommand{\arraystretch}{1.1}
  \captionsetup{justification=centering,margin=2cm}
  \hspace*{0cm} 
  \vspace{0.5cm} 
  \rotatebox{180}{\parbox{20cm}{\captionof{table}{\label{tab:dt_table} Performance of DTs tested on different scenarios. In bold are marked the scenarios that will be reported in Table \ref{tab:comp_table} for comparison purposes.}}}

\rotatebox{180}{\begin{tabular}{lllggggllllgggg}
    \hline 
    & & & \multicolumn{12}{c}{Data set} \\
    \cline{4-15}
    & & & \multicolumn{4}{c}{Noise 1} & \multicolumn{4}{c}{Noise 10} & \multicolumn{4}{c}{Noise 50} \\
    \hline
    $\begin{array}{l}
    \text {Data} \\
    \text {transformation}
    \end{array}$ & $\begin{array}{l}
    \text {Splitting} \\
    \text {criterion}
    \end{array}$ & $\begin{array}{l}
    \text {DT} \\
    \text {prun.}
    \end{array}$  & \text {Depth} & $\begin{array}{l}
    \text {\# of} \\
    \text {nodes}
    \end{array}$   & $\begin{array}{l}
    \text {Train} \\
    \text {acc.}\\
    \text {($\%$)}
    \end{array}$  & $\begin{array}{l}
    \text {Test} \\
    \text {acc.}\\
    \text {($\%$)}
    \end{array}$
    & \text {Depth} & $\begin{array}{l}
    \text {\# of} \\
    \text {nodes}
    \end{array}$   & $\begin{array}{l}
    \text {Train} \\
    \text {acc.}\\
    \text {($\%$)}
    \end{array}$  & $\begin{array}{l}
    \text {Test} \\
    \text {acc.}\\
    \text {($\%$)}
    \end{array}$
    & \text {Depth} & $\begin{array}{l}
    \text {\# of} \\
    \text {nodes}
    \end{array}$   & $\begin{array}{l}
    \text {Train} \\
    \text {acc.}\\
    \text {($\%$)}
    \end{array}$  & $\begin{array}{l}
    \text {Test} \\
    \text {acc.}\\
    \text {($\%$)}
    \end{array}$\\
    \hline
  \multirow{6}{*}{STD} & \multirow{3}{*}{Entropy} & no & 15 & 454 & 100 & 93.69 & 16 & 480 & 100 & 93.45 & 19 & 1018 & 100 & 84.39\\ 
  & & pre & 13 & 416 & 99.47 & 93.61 & 13 & 428 & 99.37 & 93.93 & 12 & 782 & 97.17 & 84.87\\ 
  & & post & 12 & 154 & 95.57 & 91.59 & 12 & 142 & 95.31 & 91.99 & 10 & 128 & 87.15 & 83.5\\ \hhline{~--------------}
  & \multirow{3}{*}{Gini} & no & 15 & 530 & 100 & 93.45 & 14 & 600 & 100 & 93.37 & 19 & 1078 & 100 & 83.25\\ 
  & & pre & 13 & 520 & 99.84 & 93.61 & 13 & 556 & 99.55 & 93.28 & 10 & 686 & 95.47 & 83.25\\ 
  & & post & 10 & 116 & 93.75 & 89.56 & 10 & 106 & 91.42 & 90.13 & 9 & 82 & 85.17 & 81.96\\
  \hline
  \multirow{6}{*}{COV} & \multirow{3}{*}{Entropy} & no & 6 & 48 & 100 & 98.28 & 8 & 54 & 100 & 98.9 & 11 & 118 & 100 & 94.99\\ 
  & & pre & 5 & 44 & 99.57 & 98.28 & 5 & 44 & 98.98 & 98.75 & 5 & 50 & 95.54 & 91.86\\ 
  & & post & 6 & 22 & 98.83 & 97.97 & 6 & 26 & 98.94 & 98.59 & 6 & 24 & 95.61 & 92.8\\ \hhline{~--------------}
  & \multirow{3}{*}{Gini} & no & 7 & 73 & 100 & 98.9 & 9 & 80 & 100 & 98.59 & 10 & 136 & 100 & 95.62\\ 
  & & pre & 6 & 60 & 99.61 & 99.06 & 6 & 62 & 99.41 & 98.28 & 6 & 76 & 97.65 & 95.77\\ 
  & & post & 6 & 26 & 98.63 & 98.44 & 5 & 24 & 98.16 & 98.28 & 7 & 32 & 97.06 & 95.15\\
  \hline
  \multirow{6}{*}{COV-PCA(4)} & \multirow{3}{*}{Entropy} & no & \textbf{9} & \textbf{48} & \textbf{100} & \textbf{99.22} & 8 & 44 & 100 & 98.75 & 26 & 792 & 100 & 78.72\\ 
  & & pre & 5 & 30 & 99.61 & 99.06 & 7 & 40 & 99.92 & 98.75 & \textbf{6} & \textbf{102} & \textbf{83.95} & \textbf{82.79}\\ 
  & & post & 4 & 12 & 99.26 & 98.75 & 4 & 14 & 98.86 & 98.9 & 10 & 66 & 83.4 & 82.0\\ \hhline{~--------------}
  & \multirow{3}{*}{Gini} & no & 9 & 50 & 100 & 98.9 & 7 & 58 & 100 & 99.37 & 20 & 820 & 100 & 77.93\\ 
  & & pre & 5 & 34 & 99.65 & 98.9 & \textbf{7} & \textbf{58} & \textbf{100} & \textbf{99.37} & 7 & 206 & 87.67 & 80.44\\ 
  & & post & 4 & 12 & 99.14 & 98.75 & 4 & 12 & 98.94 & 99.06 & 6 & 24 & 81.4 & 79.97\\
  \hline
\end{tabular}}
\end{sidewaystable}

\section{Support Vector Machine}\label{sec: svm}

Support Vector Machines (SVMs) are ML algorithms that attempt to draw a plane between binary classified data. In the original paper \cite{Boser92}, the authors first explain how an optimal hyperplane can be found. This plane can be described as
 
\begin{equation}
    D(x) = \sum_{i=0}^{N}\omega_{i}\phi_{i}(x) + b,
\end{equation}

where $x$ is the input and $\phi_i$ is a user-defined basis function. Lastly, $\omega_i$ and $b$ are the trainable weights and bias usually found by solving an optimisation problem. The binary classification of the data is based on the sign of the decision function $D(x)$.

The decision function may also be written as 
\begin{equation}
    D(x) = \sum_{j=0}^{l} y_j \alpha_{j} K(x_{j}, x) + b.
\end{equation}
Here, $\alpha_i$ and $b$ are the trainable parameters. The function $K$ is a kernel related to the user functions $\phi_i$ and $x_j$ are input data. These components are obtained from the dual of the optimisation problem referred to above. In modern software, the kernel is typically defined by the user such that the basis function is never explicitly defined. Commonly used kernels are linear, polynomial and a variety of radial basis functions (RBF).

In \cite{Boser92}, the authors demonstrate that training the ML method involves solving a convex quadratic program. The soft margin was later introduced in \cite{Cortes1995}, \rex{using $l_2$-penalization of mislabelled data points, thereby allowing for a feasible solution in the case of overlapping classes.} 
Our model is trained by solving the quadratic program that follows,

\begin{minipage}[t]{0.49\linewidth}
\vspace{1mm}
\begin{center}
    Primal
\end{center}%
\vspace{0mm}
 
\begin{align*}
    \min_{\alpha, \xi, b}& \quad \frac{1}{2}\omega^2 + C \xi^{\top}\mathbb{1}\\
    \text{s.t.}& \qquad y_i(\omega^{\top}\phi(x_i) + b) > 1 - \xi_i\\
               & \qquad \xi_i \geq 0 \\
               & \qquad \text{for all $i$}
\end{align*} 

\vspace{1mm}
\end{minipage}%
\begin{minipage}[t]{0.49\linewidth}
\vspace{1mm}
\begin{center}
    Dual
\end{center}%
\vspace{0mm}
\begin{align*}
    \min_{\alpha}& \quad \frac{1}{2} \alpha^{\top} H \alpha  - \alpha^{\top}\mathbb{1}\\
    \text{s.t.}& \qquad \alpha^{\top} Y = 0\\
               & \qquad 0 \leq \alpha \leq C \mathbb{1},
\end{align*} 
\vspace{1mm}
\end{minipage}
\rex{and differs slightly from the original method in \cite{Boser92} as it uses $l_1$-penalization of mislabelled data.} Here $Y = \{y_0, \dots, y_p\}$ are the classifications of the data set, $H$ is an $l\times l$ matrix with elements $H_{ij} = y_i y_j K(x_i, x_j)$. The hyperparameter $C$ allows for a soft margin and $\xi_i$ is the measure of the deviation of point $x_i$ from the margin. Any data point $x_i$ for which the corresponding $\alpha_i > 0$ is considered a support vector. Penalizing the deviations by increasing $C$ increases the number of support vectors, which may lead to overfitting. 

\subsection{Experiments}

In this subsection, we evaluate the performance of the SVM through a set of experiments. As in the previous two sections, we apply a dispersion method to transform the data. When the transformation involves the covariance matrix we have also, for comparability between transformations, applied SVM to the top three PCs. 

\begin{figure}
    \centering
     \begin{subfigure}[b]{0.49\linewidth}
      \centering
        \includegraphics[trim={0 0 19.5cm 0},clip, width=\linewidth]{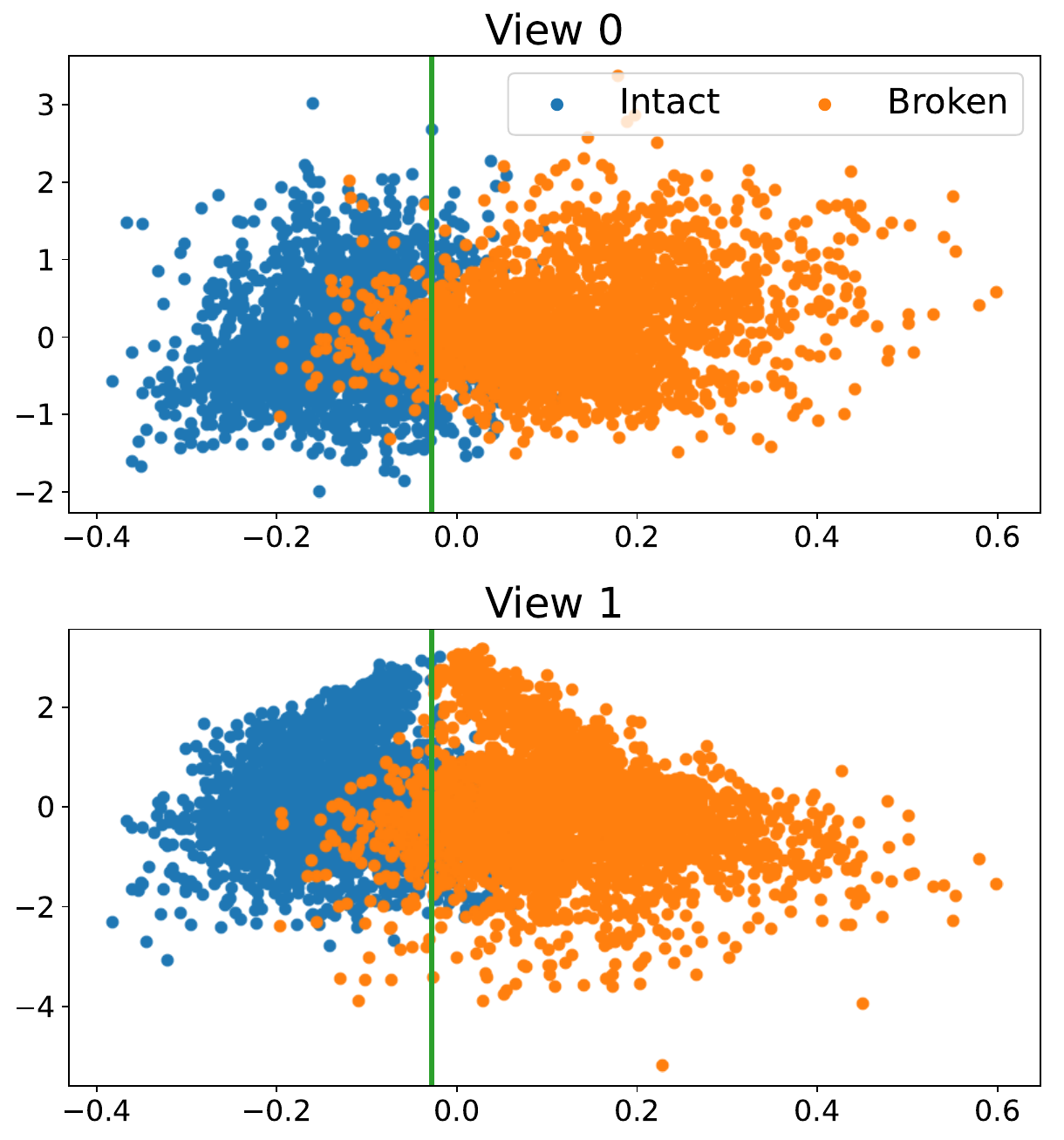}
     \end{subfigure}
     \begin{subfigure}[b]{0.49\linewidth}
     \centering
        \includegraphics[trim={0 0 19.5cm 0},clip, width=\linewidth]{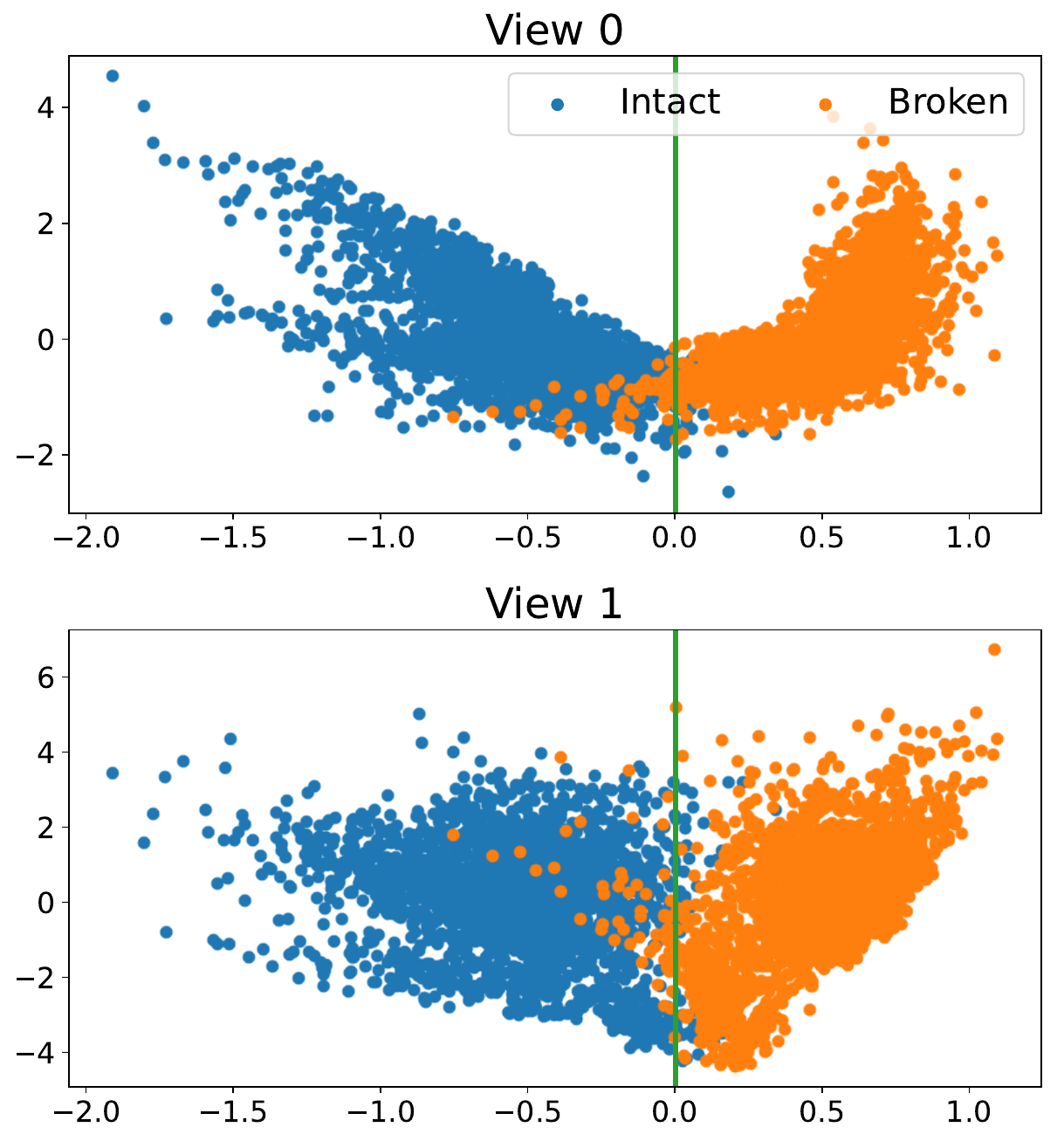}
     \end{subfigure}
     \caption{Figure showing linear SVMs performance on dataset with STD transform  (left column) or  COV transform and 3 PCs (right column). Both are created from a subset of the data set containing only one physical direction.}
      \label{fig:SVM Slack projected}
\end{figure}

For experiments limited to three dimensions the results are visualised in Figure \ref{fig:SVM Slack projected}. The plots illustrate how a linear plane is able to separate the data points. One can see how the data is relative to the decision border of the linear SVM both for STD transform and COV transform with 3 PCs. 

In the experiments, SVMs are trained with either an RBF or a linear kernel. For each choice of kernel, every combination of number of PCs, transformation method and noise level is tested. For each test, the hyperparameter $C$ is optimised using \texttt{sklearn}s \texttt{GridSearchCV} method. The test accuracy is reported in Table \ref{tab:SVM} along with the number of support vectors needed by the RBF SVMs. For the linear SVM the hyperplane is defined by $n+1$ coefficients, where $n$ is the number of PCs.

Although there is overlap between all the point clouds in Figure \ref{fig:SVM Slack projected}, the PCA based model manages a greater relative distance to the hyperplane, indicating higher robustness. 
This also becomes apparent by inspecting the number of support vectors for the cases with the same number of PCs, but different transformations, in Table \ref{tab:SVM}. The STD based approaches need significantly more support vectors than the COV based, while still performing worse on the test set. SVMs using the COV transform and $7$ PCs, essentially spanning the whole data set, only needed a few more support vectors than the ones with $21$ PCs. Given that the SVM with RBF kernel relies on a number of support vectors much larger than the number of PCs, it is slower to evaluate than the linear SVM.

\newcolumntype{h}{>{\columncolor{mygray}}r}

\begin{table}
\begin{tabular}{lh|hh|r|rr|h|hh}
\hhline{----------}
\multirow{3}{*}{$
 \begin{array}{l}
      \text{Data}\\
      \text{trans.}\\
      \text{(\#PCs)}
 \end{array}$}
 & \multicolumn{3}{c}{Noise 1} & \multicolumn{3}{c}{Noise 10} & \multicolumn{3}{c}{Noise 50} \\
\hhline{~---------}
 & \multicolumn{1}{c}{Linear} \vline & \multicolumn{2}{c}{RBF} \vline& \multicolumn{1}{c}{Linear} \vline& \multicolumn{2}{c}{RBF} \vline& \multicolumn{1}{c}{Linear} \vline& \multicolumn{2}{c}{RBF} \\
\hhline{~---------}
 & Acc. & Acc. & SV & Acc. & Acc. & SV & Acc. & Acc. & SV \\
\hhline{----------}
STD(3)* & 0.940 & 0.950 & 1264 & 0.866 & 0.874 & 1866 & 0.650 & 0.668 & 3591 \\
COV(3)* & 0.986 & 0.986 & 465 & 0.974 & 0.987 & 568 & 0.928 & 0.923 & 1066 \\
COV(4)* & 0.983 & \textbf{0.990} & 418 & \textbf{0.988} & 0.984 & 441 & 0.927 & \textbf{0.940} & 980 \\
COV(6)* & 0.994 & 0.999 & 364 & 0.983 & 0.994 & 444 & 0.933 & 0.942 & 954 \\
STD(6) & 0.978 & 0.983 & 969 & 0.926 & 0.942 & 1345 & 0.682 & 0.726 & 3239 \\
COV(6) & 0.988 & 0.993 & 621 & 0.982 & 0.994 & 616 & 0.946 & 0.958 & 992 \\
COV(7) & 0.993 & 0.998 & 484 & 0.993 & 0.996 & 481 & 0.953 & 0.970 & 853 \\
COV(21) & 0.999 & 1.000 & 462 & 0.996 & 0.998 & 519 & 0.947 & 0.972 & 923 \\
\hhline{----------}
\end{tabular}
\caption{
\rez{Accuracy for linear SVM and RBF SVM applied to the noisy test sets. The number of support vectors for the SVM with the RBF kernel is given in the SV columns.  An asterisk (*) indicates that only one physical direction was used from the sensors. In bold are marked the scenarios that will be reported in Table \ref{tab:comp_table} for comparison purposes.}
}
\label{tab:SVM}
\end{table}

\section{Convolutional Neural Networks}\label{sec: cnn}
As \rex{mentioned} in Section \ref{sec: dataset}, the supervised learning task consists of estimating the function $F$ in \eqref{eq:Fsp} through a parameterized function $F_\theta$, with $\theta$ representing the parameters \rex{to be learnt}. In this section, we illustrate how neural networks can provide a useful framework to achieve this task.

In the most basic form of fully connected, feedforward neural networks, the input-output mapping $F_{\theta}$ is obtained by a composition of nonlinear functions $\phi$:
 
\begin{equation} \label{eq:Ftheta}
F_{\theta}(x_0)=\phi_L \circ \phi_{L-1} \circ \cdots \circ \phi_l \circ \ldots \phi_1 (x_0),
\end{equation}

with $x_0\in\mathbb{R}^{n_0}$ a given input data, $L$ the number of layers in the network\rex{, which determines its depth, and $\phi_l: \mathbb{R}^{n_{l-1}}\to \mathbb{R}^{n_l}$, $\phi_l(x_{l-1}):=\sigma\left(W^lx_{l-1}+b^l\right)$ for $l=1,  \ldots, L$.} We also refer to these networks as multilayer perceptrons (MLPs). \rex{Weight matrices $W^{l} \in \mathbb{R}^{n_l \times n_{l-1}}$ and bias vectors $b^{l} \in \mathbb{R}^{n_l}$ contain trainable parameters.} 
The nonlinear activation function $\sigma: \mathbb{R}^{n_l} \rightarrow \mathbb{R}^{n_l}$, acting component-wise, typically belongs to $C^0$ and is monotonically non-decreasing. Examples of such functions are the sigmoid function and the \rez{rectified linear unit (ReLU)}. \rex{The training procedure consists of minimising a differentiable loss function, that quantifies the discrepancy between the predictions of the network and the labels, over the network parameters. Usually, a stochastic gradient descent algorithm is used.}

Convolutional Neural Networks (CNNs) use particular affine mappings in the feedforward propagation of the input data. In the following, we consider one-dimensional CNNs, where each layer applies a one-dimensional linear kernel $K$ over sections $S$ of the input data, to detect relevant features. Assuming that both the filter $K$ and the receptive field $S$ are defined on the integer $i$, with $S\in \mathbb{R}^s$ and $K$ having finite support in the set $\{1-s,2-s,\dots,s-2,s-1\}$, this operation corresponds to a discrete convolution
\begin{equation*}
(S * K)(i)=\sum_{j=1}^s S(j) K(i-j).
\end{equation*}
The parameters to be determined during the training are the entries of the linear filters. This results in a significant reduction in parameters, in contrast to \rex{dense} fully connected neural networks.
\rey{It should be noted that, reflecting the filter, the convolution operation can be interchanged with correlation. Therefore, since the filter is learnable, its application can also be described in terms of correlation. 
Input data can include multiple channels, which may vary across different layers. In such cases, the filters are represented by tensors and the convolution operation becomes multidimensional. This allows for the learning of unique features for each channel and the generation diverse feature maps.} 
Each convolutional layer is followed by a pooling layer which uses pooling filters to reduce the dimensionality of the feature maps. The most commonly used pooling techniques are max pooling and average pooling, which, respectively, propagate the maximum and average values from sections of the feature maps \cite{higham2019}.

As a result, we can model the forward propagation of the input data in a CNN as a composition of mappings $\phi_{cn}$ given by
\begin{equation*}
  \phi_{cn}: \mathbb{R}^{{n_{l-1}}\times m_{l-1}} \to \mathbb{R}^{{n_{l}}\times m_{l}},\qquad \phi_{cn}(x_{l-1}) = P(\sigma(C(x_{l-1}))),  
\end{equation*}
where $n_l$ and $m_l$ are, respectively, the length and the number of channels of the output tensor of layer $l$, $C$ is a convolution operator resulting from sliding linear filters across the feature maps from the previous layer and adding a bias, $\sigma$ is a nonlinear activation function, and $P$ is a pooling operator that coarsens the grid over which the feature maps are defined \cite{bronstein2021geometric}. 
Moving deeper into the network, higher-level features are created. The ones returned from the final pooling layer are usually mapped to a vector and fed to an MLP, which returns a prediction about the class label.

\begin{figure}[htbp]
\includegraphics[width=\textwidth]{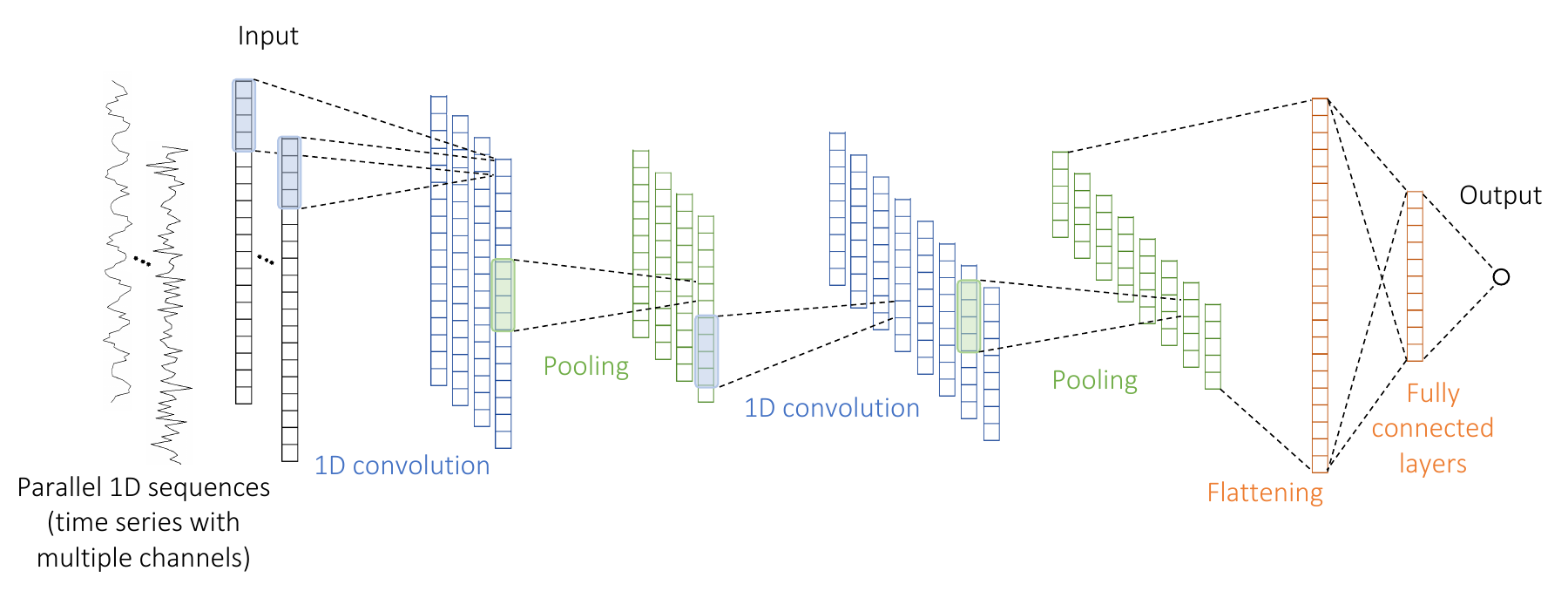}
\caption{A typical one-dimensional CNN architecture.}
\label{fig:cnn_architectur}
\end{figure}

\subsection{Experiments}

\begin{figure}[htbp]
\centering
\includegraphics[width=0.32\textwidth]{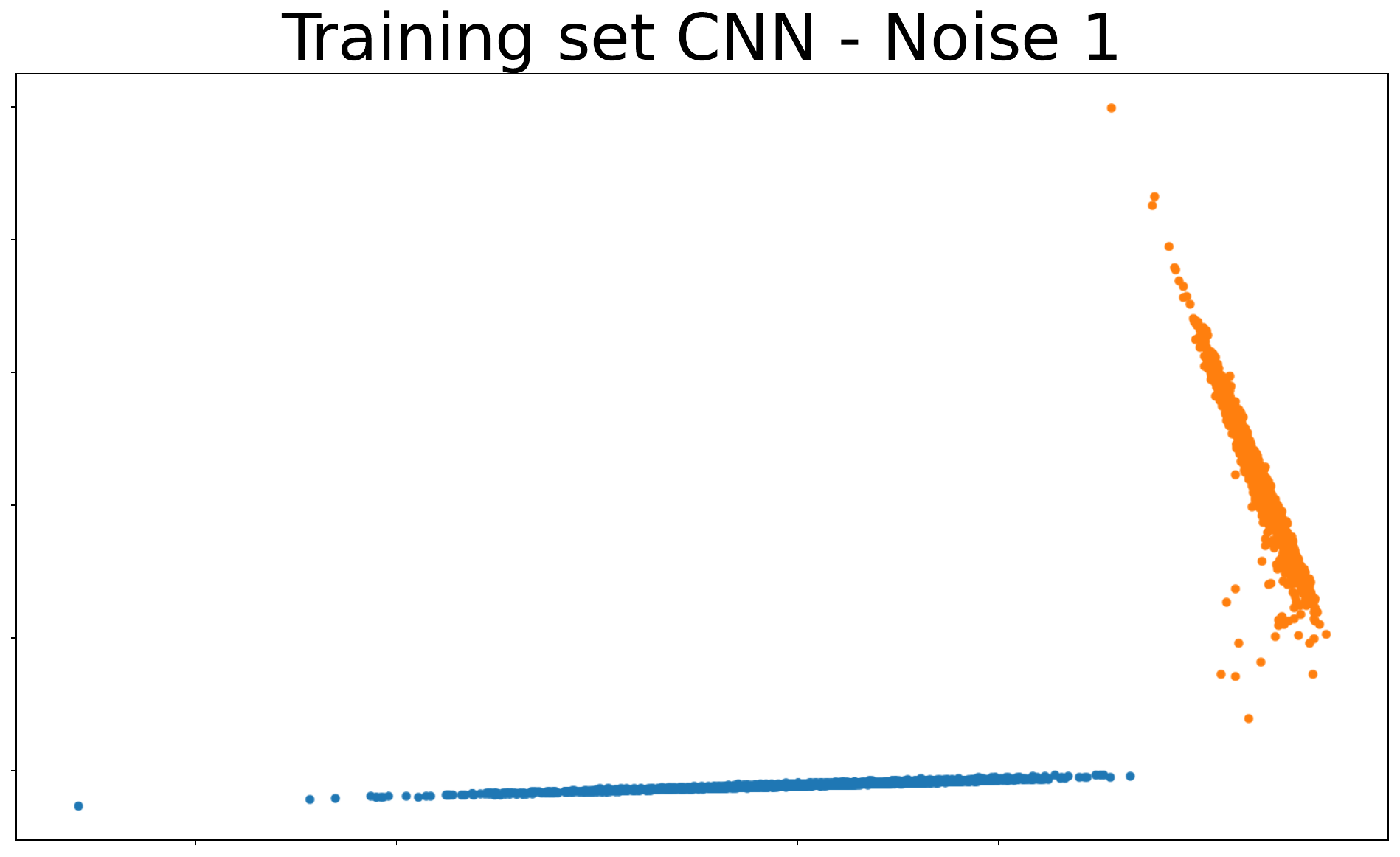}
\includegraphics[width=0.32\textwidth]{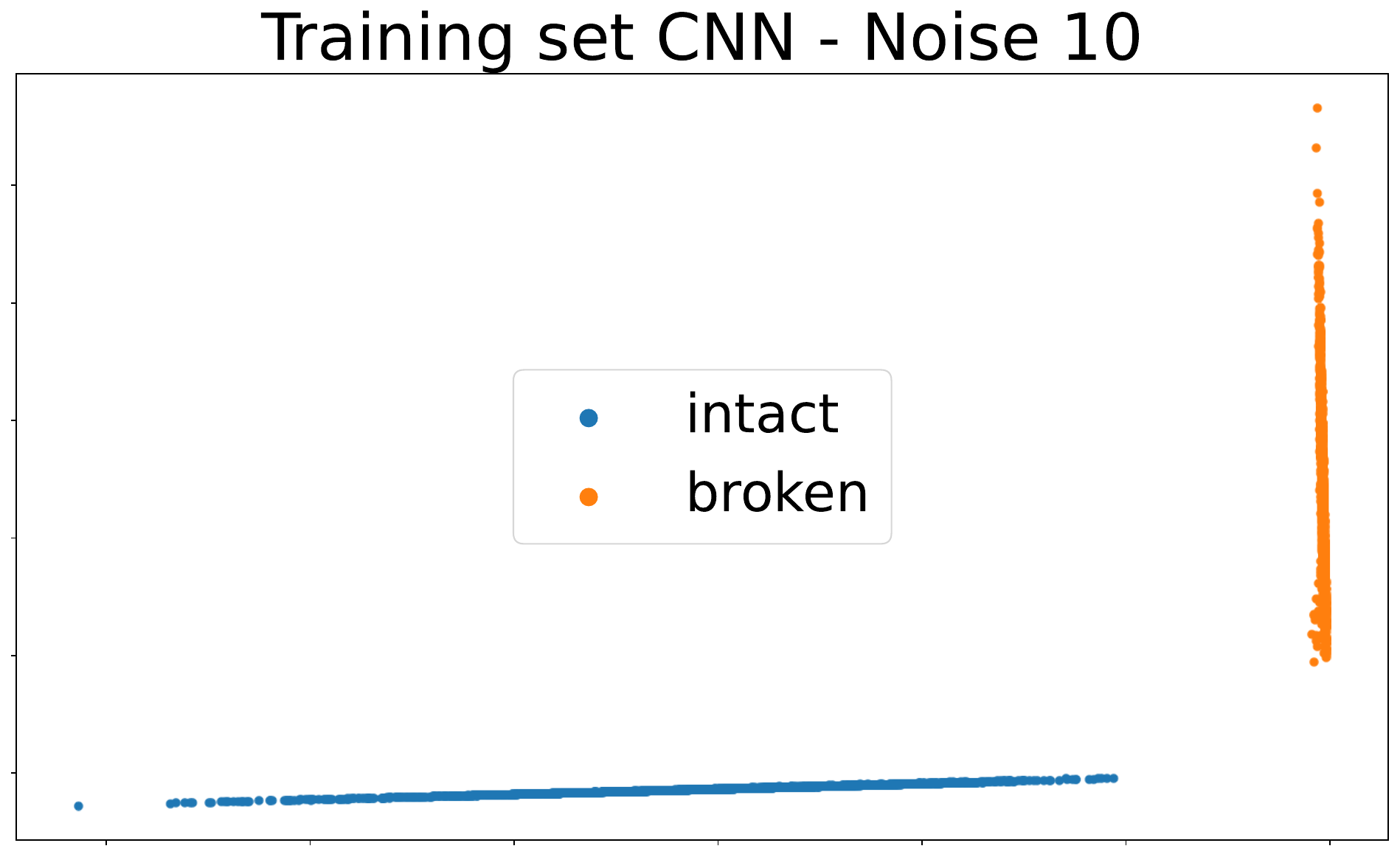}
\includegraphics[width=0.32\textwidth]{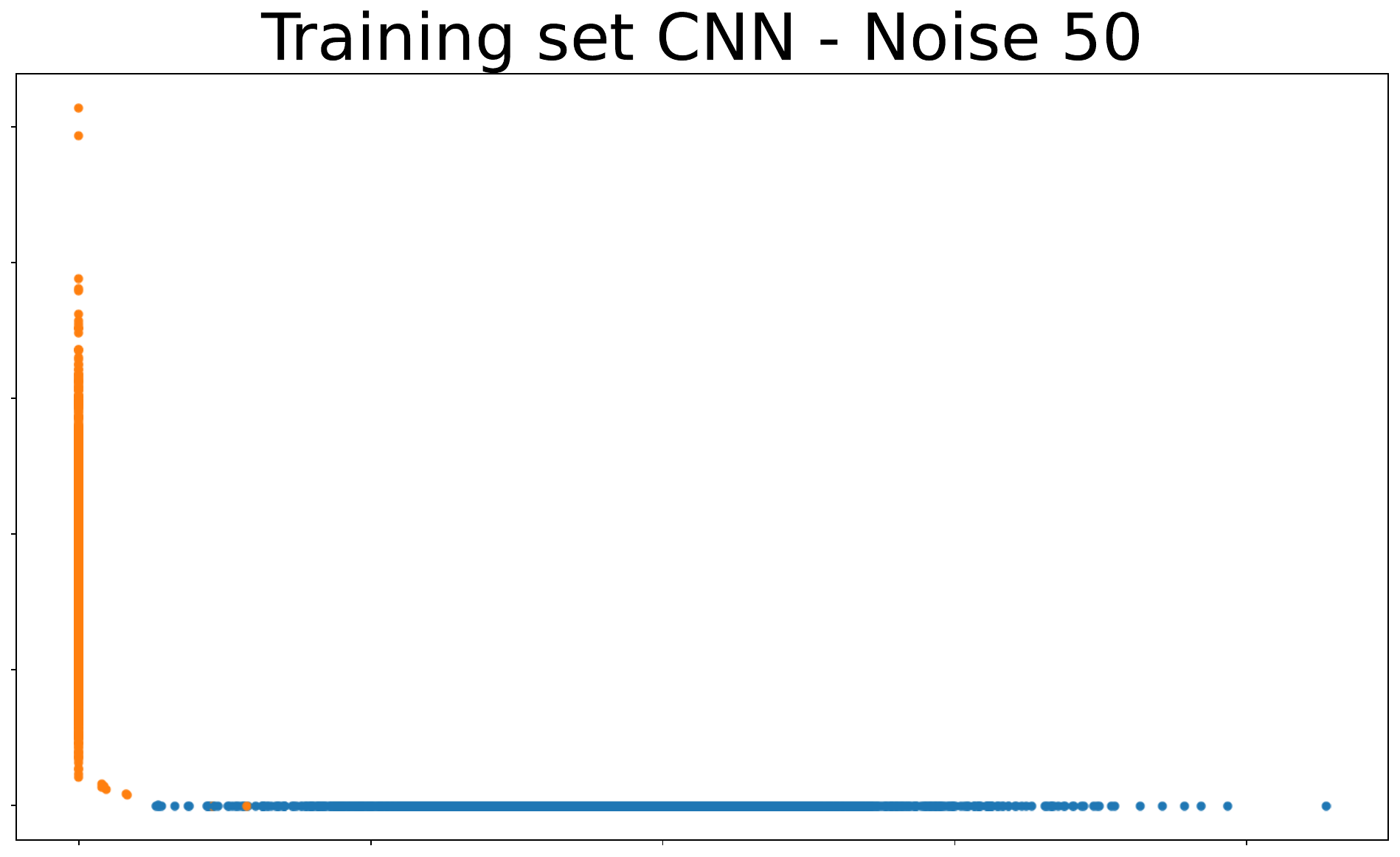}
\includegraphics[width=0.32\textwidth]{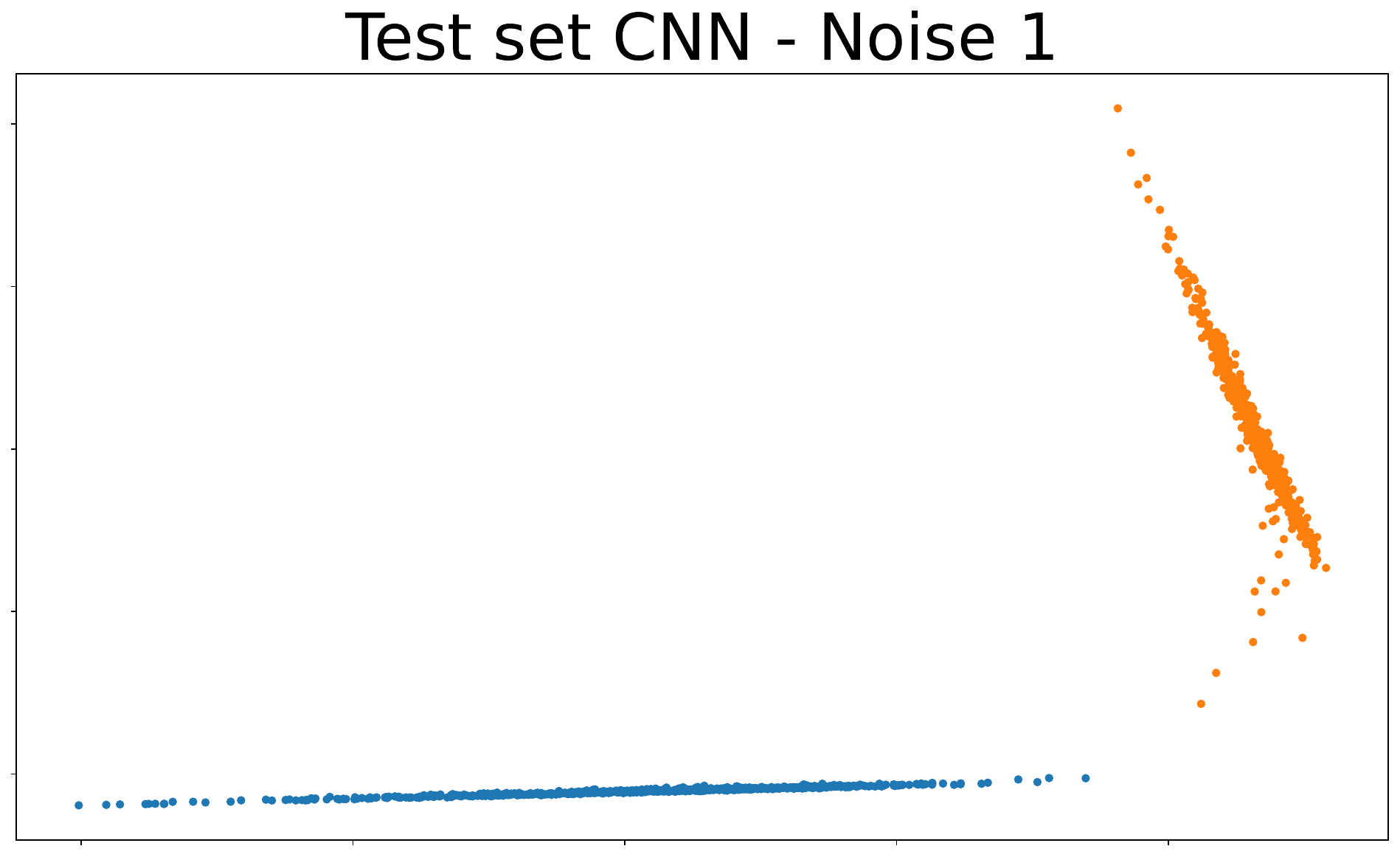}
\includegraphics[width=0.32\textwidth]{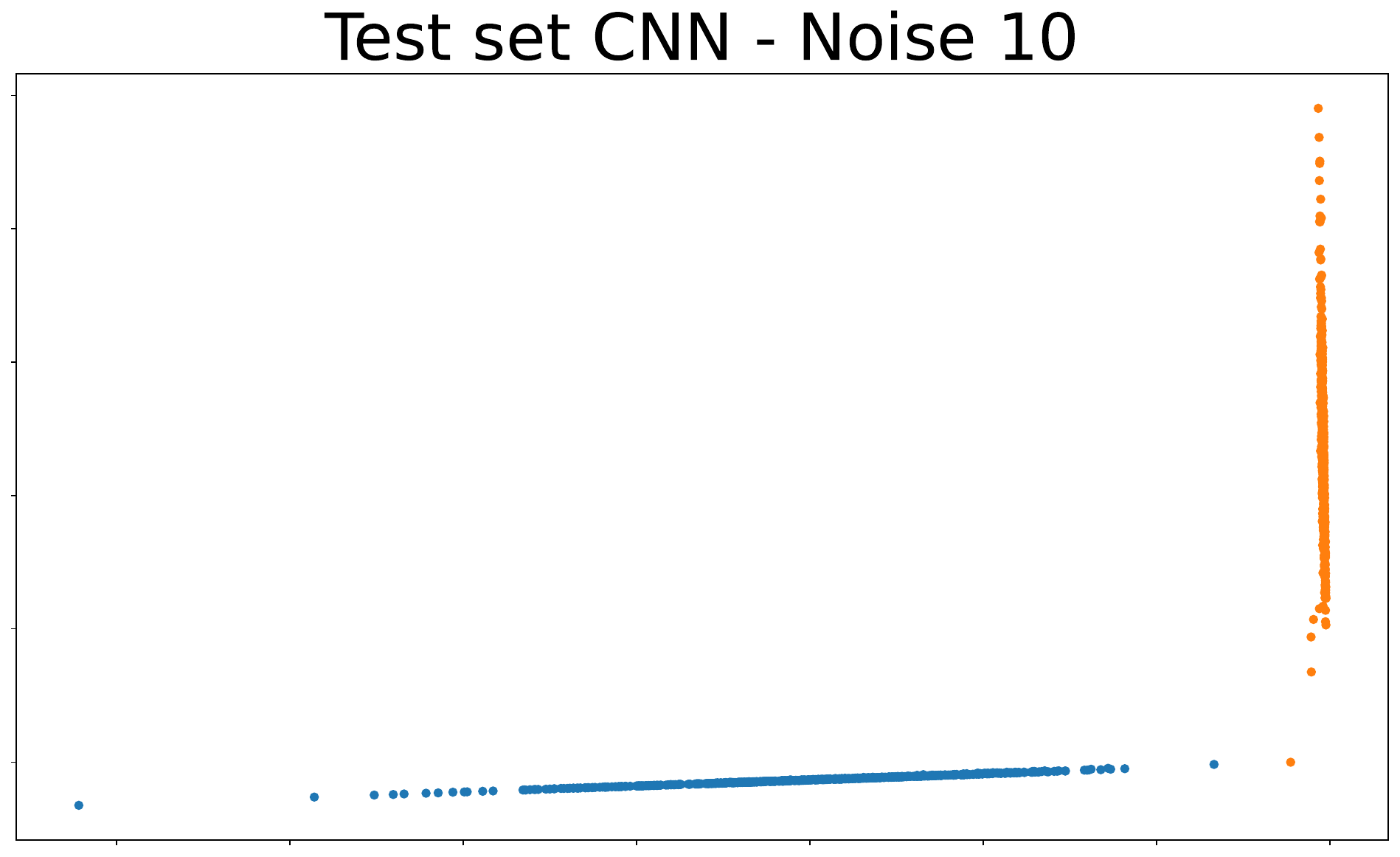}
\includegraphics[width=0.32\textwidth]{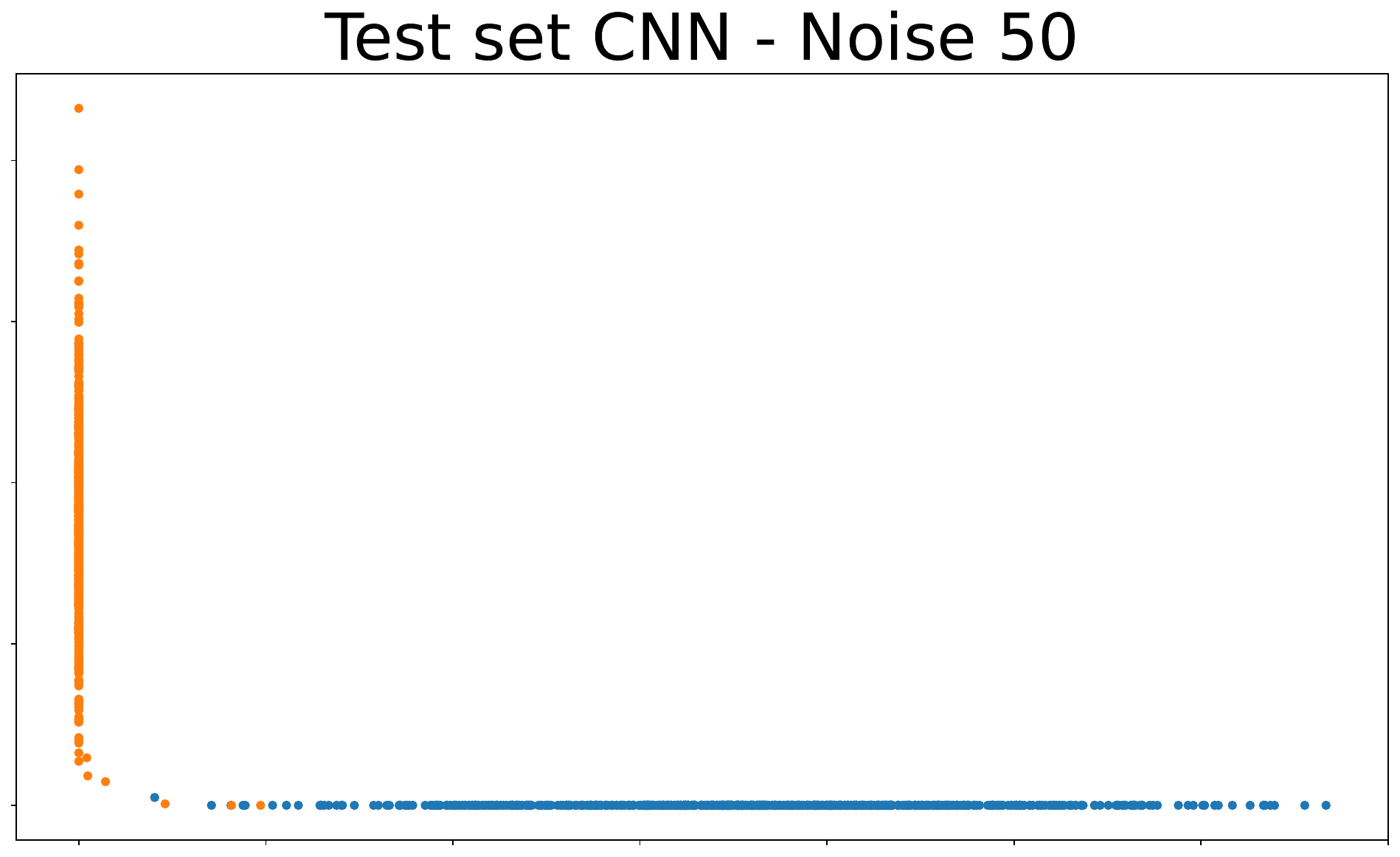}
\caption{\rez{The figures illustrate the transformation of the input data by the CNN in both the training and test sets, under the three different noise scenarios. Prior to the output layer, which predicts the class, each individual time series is converted into a two-dimensional vector and can be visually represented as a point on a plane. In the case of Noise 1 and Noise 10, the data points belonging to the two categories form separate clusters.}}
\label{fig:cnn_transform}
\end{figure}

\rex{The time series in the original data set were split into \rex{one-minute intervals and collected into non-overlapping training and test sets, with the former containing 80\% of the resulting series and the latter the remaining 20\%.} In Figure \ref{fig:cnn_transform}, we show the results obtained using a CNN with 3 convolutional layers, each of which doubles the number of channels and is followed by an averaging pooling layer. \rez{Finally, an MLP consisting of one hidden layer and an output layer consisting of a sigmoid function is used for prediction}. To assign a label to the input data, a threshold is fixed to $0.5$, so that when the output is greater than or equal to the threshold, the input time series is classified as broken, or intact otherwise. Details on network architecture can be found in the code snippet listed in Appendix \ref{sec:supp_mat_cnn}, written in \texttt{PyTorch} \cite{paszke2019pytorch}.}

\rex{The experiments are run with the number of epochs set to 100. The activation function and certain hyperparameters in the training procedure are varied using the \texttt{Optuna} software framework \cite{optuna_2019}. More specifically, we evaluate different values of batch size, learning rate, and weight decay for the Adam algorithm \cite{Adam_KingBa15}, which is used as optimiser. The specific ranges for each parameter are listed in Table~\ref{tab:hyperparams_cnn} in the Appendix. The loss function is defined as the mean squared error (MSE) between the true labels and the predictions of the network. The combinations of hyperparameters yielding the best results on the test set for each level of noise, along with the corresponding mean squared errors on the training and test sets, are presented in Table \ref{tab:best_hyperparams_cnn}.}

\begin{table}[htbp]
\centering
\small
{
\begin{tabularx}{1.\textwidth}{ 
   >{\centering\arraybackslash}X 
   >{\centering\arraybackslash}X
   >{\centering\arraybackslash}X
   >{\centering\arraybackslash}X}
\toprule
\multicolumn{4}{c }{Selected hyperparameters}\\
\midrule
        & Noise 1 & Noise 10 & Noise 50 \\
\midrule
     activation function & LeakyReLU & LeakyReLU  & Swish \\
     learning rate $\eta$ & $2.562 \cdot 10^{-2}$ &  $2.102 \cdot 10^{-3}$  & $1.017 \cdot 10^{-2}$ \\
     weight decay & $1.243 \cdot 10^{-5}$  & $1.221 \cdot 10^{-5}$   &  $1.520 \cdot 10^{-7}$ \\
     batch size & 30 & 10  & 30 \\
\midrule
MSE train & $8.856 \cdot 10^{-6}$ & $5.968 \cdot 10^{-5}$ & $6.068 \cdot 10^{-4}$ \\
MSE test & $2.815 \cdot 10^{-5}$ & $3.054 \cdot 10^{-4}$ & $2.427 \cdot 10^{-3}$\\
 \bottomrule
\end{tabularx}
}
\caption{Combination of hyperparameters yielding the best results in each scenario, corresponding to the plots in Figure \ref{fig:cnn_transform}, after conducting 100 trials with \texttt{Optuna}.}
\label{tab:best_hyperparams_cnn}
\end{table}

\section{Comparison of methods}\label{sec:comparison}

\begin{figure}[htbp]
    \centering
    \includegraphics[width = 0.85\textwidth]{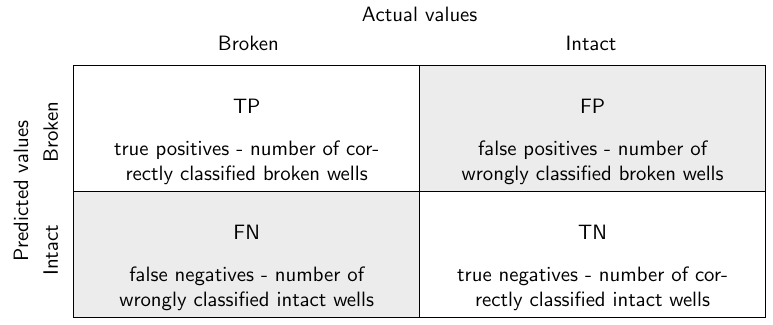}
    \caption{Confusion matrix used to evaluate the performance of the classification techniques.}
    \label{fig:confusion_matrix_broken_intact_narrow}
\end{figure}

In this section, we compare the tested methods based on performance metrics. We consider precision, recall and F1-score, defined in terms of the entries in the so-called confusion matrix in Figure \ref{fig:confusion_matrix_broken_intact_narrow} as:
\[
\text{Precision} = \frac{TP}{TP+FP}, \quad \text{Recall} = \frac{TP}{TP+FN} , \quad \text{F1-score} = \frac{2\cdot \text{Precision} \cdot \text{Recall}}{\text{Precision} + \text{Recall}}.
\]
In Table \ref{tab:comp_table}, we report the performance of the methods measured with the \texttt{Python} functions of \texttt{sklearn.metrics}: \texttt{classification\_report} gives  the precision, recall and F1 scores.

For the methods where we have tested different scenarios, we report here only the best-performing ones, \rez{marked in bold in the respective sections}. \rex{The results indicate that all the classical ML algorithms perform similarly well in terms of accuracy, but are outperformed by the more advanced CNN. For the different methods there are significant differences in the train and test times.}

Already with 4 PCs, LogR-PCA  shows almost perfect results. The number of parameters needed to make the classifications is only one more than the dimensionality of the data, proving that non-complex algorithms could suffice in classification of the data. The decision trees score second to best using 4 PCs, but needs significantly more parameters than the LogR. As the dimensionality increases so does the number of parameters, making it prone to overfitting. The SVM gets a lower comparative score than the two previously mentioned methods, and needs 940 support vectors. However, as the number of PCs increases, the number of support vectors is reduced, as seen in Section \ref{sec: svm}. This suggests that the SVM would perform better and with higher robustness on a data set with increased dimensionality than e.g. the DTs. Finally, \rex{CNN provides the best results in terms of accuracy, and is able to correctly classify all the time series in the Noise 1 and Noise 10 datasets}, without requiring pre-processing with PCA and COV-transform. As is common for deep learning algorithms, however, it requires longer \rex{offline} training time, and a fine tuning of different hyper-parameters.

\begin{table}[htbp]
\setlength{\tabcolsep}{3.2pt}
\renewcommand{\arraystretch}{1.1}
\begin{tabular}{lllllll}
    \hline 
    \text {Data set} & \text {Method} & \text{Precision} & \text {Recall} & $\begin{array}{c}
    \text {F1} \\
    \text {Score}
    \end{array}$ & $\begin{array}{c}
    \text {Train} \\
    \text {Time} \\
    \text {(ms)}
    \end{array}$ & $\begin{array}{c}
    \text {Test} \\
    \text {Time} \\
    \text {(ms)}
    \end{array}$ \\
    \hline \multirow{4}{*}{Noise 1} & \text {LogR-PCA} & 0.997 & 0.997 & 0.997 & 10.195 & \textbf{0.990} \\
    & \text {DT-PCA} & 0.997 & 0.987 & 0.992 & \textbf{~6.662} & 0.998\\
    & \text {SVM-PCA} & 0.990 &    0.990 &      0.990 &      133.799 &     51.615 \\
    & \text {CNN} & \textbf{1.000} & \textbf{1.000}  & \textbf{1.000} & $\sim$ 3 min & 30.535 \\
    \hline
    \multirow{4}{*}{Noise 10} & \text {LogR-PCA} & 0.997 & 0.994 & 0.995 & 12.408 & 1.001 \\
    & \text {DT-PCA} & \textbf{1.000} & 0.987 & 0.993 & \textbf{5.207} & \textbf{0.999} \\
    & \text {SVM-PCA} & 0.988 &   0.988 &     0.988 & 24.639 &      3.003 \\
    & \text {CNN} & \textbf{1.000} & \textbf{1.000} & \textbf{1.000} & $\sim$ 3 min & 27.133 \\
    \hline
    \multirow{4}{*}{Noise 50}  & \text {LogR-PCA} & 0.808 & 0.750 & 0.778 &  11.026 & 1.016 \\
    & \text {DT-PCA} & 0.830 & 0.808 & 0.819 & \textbf{10.910} & \textbf{0.994} \\
    & \text {SVM-PCA} &  0.940 &    0.940 &      0.940 & 212.493 &    106.985 \\
    & \text {CNN} & \textbf{0.995} & \textbf{1.000} & \textbf{0.998} &  $\sim$ 4 min  & 49.181 \\
    \hline
\end{tabular}\caption{\label{tab:comp_table}Performance of the methods. Given the high scoring of the classical ML algorithms on the full data set they are here compared using $4$ PCs of the COV-transformed data set.}
\end{table}

\section{Conclusion}\label{sec:conclusion}

We observed in Section \ref{sec: dataset} that measures of statistical dispersion applied to shorter time series are a good preprocessing tool for ML algorithms not specifically designed to handle temporal dependencies. Additionally, we observed how the dimensionality of the COV-transform data set could be significantly reduced using PCA.

We presented \rey{in Section \ref{sec:SotA}} a baseline method for classifying the time series and discussed its efficiency. Given the method's reliance on human assistance, we were unable to evaluate its performance. However, we found that the method, to a certain extent, would be able to distinguish between broken or intact. Although the method is based on known statistical properties and visualization techniques, making it easy to use for practitioners, it is prone to human error.

In Sections \rey{\ref{sec:log_reg}-\ref{sec: cnn}}, popular ML algorithms were trained on the preprocessed data set. The tests showed that they performed remarkably well. In particular, the good results obtained with a simple and popular method like LogR validates the data transformation in the preprocessing phase. It was observed that the performance of SVM deteriorated faster than LogR and DTs as the dimensionality, i.e., the number of principal components, was reduced. However, the low number of support vectors needed by the SVM with sufficiently high dimensionality makes it a viable choice.

Our findings indicate that \rex{classical} ML algorithms, even when they are not \rey{originally} designed to take temporal dependencies into account, can excel in TSC given proper pre-processing. CNNs, on the other hand, suggest that deep learning is a powerful tool to extract discriminative features in time series, without the need of any data manipulation other than normalization. However, a common downside of deep learning algorithms is that the learned features do not have an immediate interpretation. \rex{Additionally, when choosing an ML method to be used in production one must carefully weigh the need for computational power versus accuracy.}

Given the experimental results, we conclude that ML algorithms are advantageous in order to reduce dependence on human decision making. In future work, it would be of interest to investigate the use of both one-class ML and unsupervised ML algorithms trained on field-measured data, as there are, to the author's knowledge, no documented measurements of a broken well. Such algorithms could be, among others, one-class SVM \cite{Scholkopf1999}, autoencoders \cite{bank2021}, CNN with Long Short Term Memory algorithms \cite{sindre2023} or isolation forests \cite{Liu2008}, which have shown good results for anomaly detection.






\section*{Acknowledgements}
The authors are grateful to Elena Celledoni, Brynjulf Owren and Mathias Hansen for the valuable discussions in various stages of this work. E.Ç. and A.L. would like to thank the group of Structural Analysis Engineering at TechnipFMC Lysaker and Kongsberg for their support and hospitality during their industrial secondment. 

This work was supported by the European Union’s Horizon 2020
research and innovation programme under the Marie Skłodowska-Curie grant agreement No. 860124. This publication reflects only the author’s view and the Research Executive Agency is not responsible for any use that may be made of the information it contains.

\clearpage 
\appendix
\section{Data set}\label{sec:appendix_data_set}

In the given maintenance operations, referenced in Section \ref{sec: intro}, the BOP is monitored through the use of Deep Water Strain sensors (DWS) and Subsea Motion Units (SMU). The DWSs give strain values at a cross-section close to the well, which again may be used to calculate loads. The SMUs are used to measure accelerations and rotational velocities above and below the flex joint that connects the riser to the BOP. In certain cases, a load relief system may be applied. One of these is the Wire Load Relief (WLR), which consists of attaching wires to the BOP and securing it to a nearby sturdy structure. Whenever WLR is used, one may also get access to the loads on each wire, but we assume that we do not in this project. 

A challenge in this project is that there exist no measurements of a well with a confirmed crack. We  model several different cases with an intact and a broken well and analyze the data. The model is set up in the commercial software Orcaflex \cite{orcaflex}. 
The data set we work with is simulated based on a generic well in the North Sea. 

When accessing a well, a decision must be made about which tools and configurations to use. This is planned before the start of each operation. Whether one or more configurations will be used varies depending on the operation being carried out. There are, however, specific configurations that, once selected, cannot be changed easily. We set up the data set as follows.

We first consider a realistic combination of permanent configurations based on
\begin{itemize}
    \item load relief (3 settings),
    \item drilling or completion (2 settings),
    \item slack or tight wellhead housing (2 settings).
\end{itemize}

Other configurations may vary. In our case, we look into
\begin{itemize}
    \item drillpipe tension (3 settings),
    \item sea states (18 settings).
\end{itemize}

\begin{figure}[htbp]
    \centering
    \resizebox{\linewidth}{!}{
        \begin{forest}
        for tree={rectangle, rounded corners,top color=white, bottom color=MatplotBlue!20,draw}
        [[NoWLR , top color=white, bottom color=MatplotRed!20 [XT [Slack [94]]
        [Tight [108]]]
        [BOP, top color=white, bottom color=MatplotRed!20  [Slack , top color=white, bottom color=MatplotRed!20  [103, top color=white, bottom color=MatplotRed!20 ]]
        [Tight [108]]]]
        [WLR-1 [XT [Slack [108]]
        [Tight [108]]]
        [BOP [Slack [108]]
        [Tight [108]]]]
        [WLR-2 [XT [Tight [46]]]
        [BOP [Tight [96]]]]]
        \end{forest}        
    }
    \caption{Number of analyses for fixed configurations. In red is the combination of configurations that we analyze in this work.}
    \label{fig:analysis_dataset}
\end{figure}
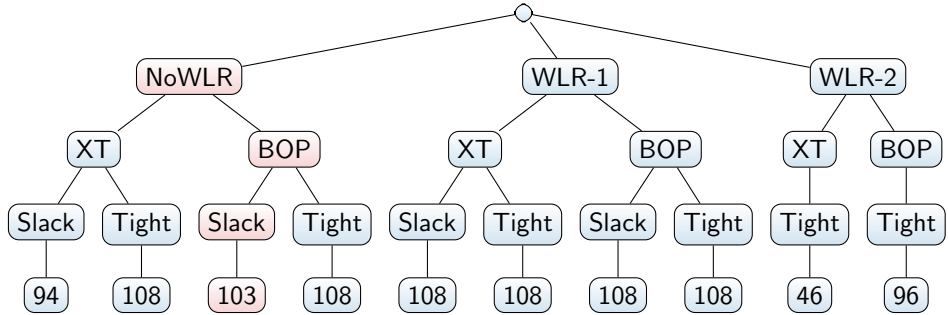

Finally, for each combination of the above configurations, two simulations are run with either the well broken or intact.
Some settings do not combine and some analyses are not able to converge, hence a total of $987$ different analyses are generated, each one hour long. Figure \ref{fig:analysis_dataset} gives an overview of the structure of the data set. 

For each analysis, three sensors are simulated at likely sensor positions. Two of these sensors, known as subsea motion units (SMUs), measure acceleration. One sensor measures strains at the wellhead and calculates bending moments, and is known as a deep water strain sensor (DWS). All of these sensors give information about the x- and y-direction and are logging at 5 Hz. A possible setup is shown in Figure \ref{fig:Stack with sensors}.


\begin{figure}[htbp]
     \centering
     \begin{subfigure}{0.59\textwidth}
        \centering
      \includegraphics[width=\linewidth]{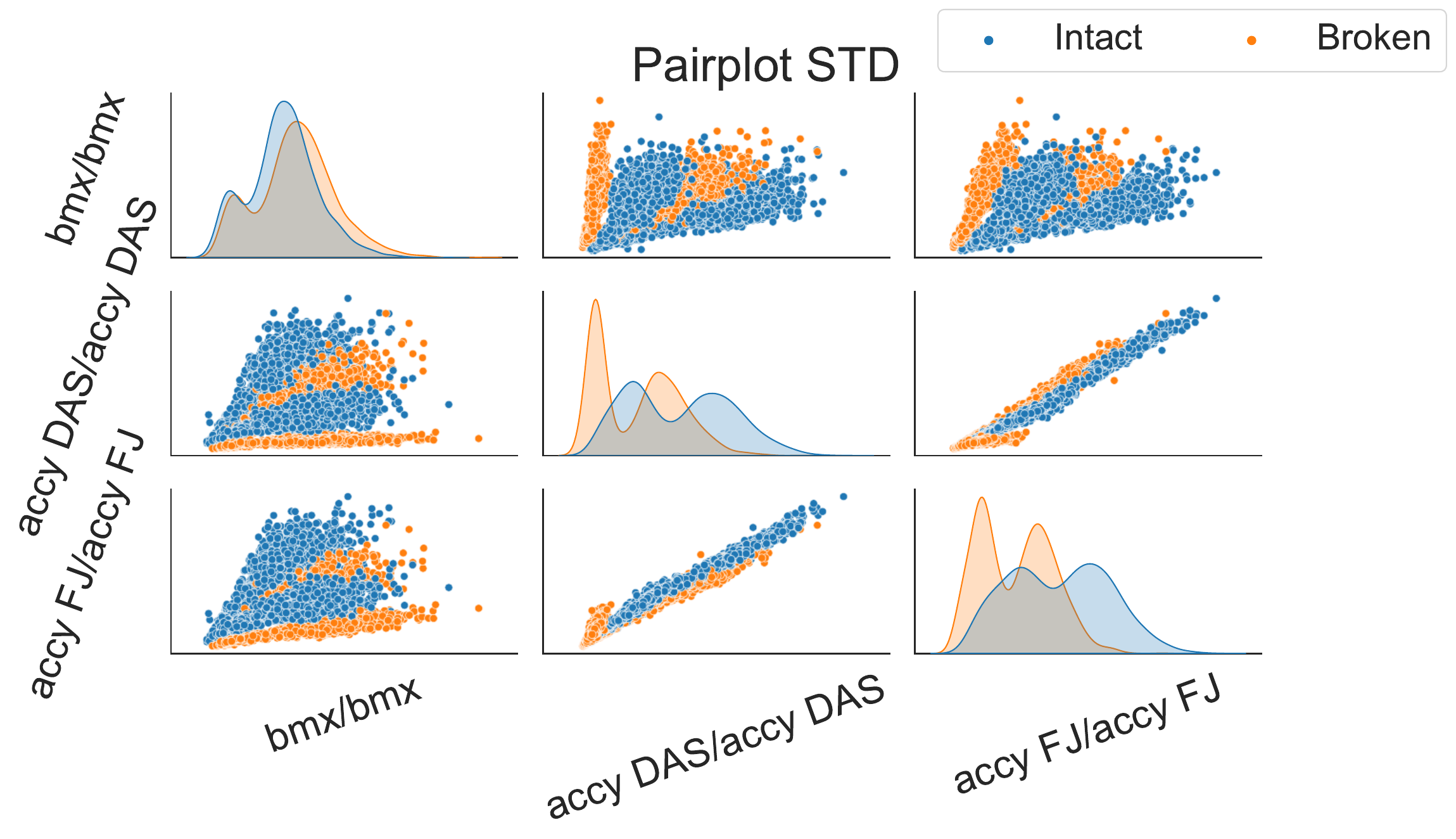}
     \end{subfigure}
     \hfill
     \begin{subfigure}{0.40\textwidth}
        \centering
      \includegraphics[width=\linewidth]{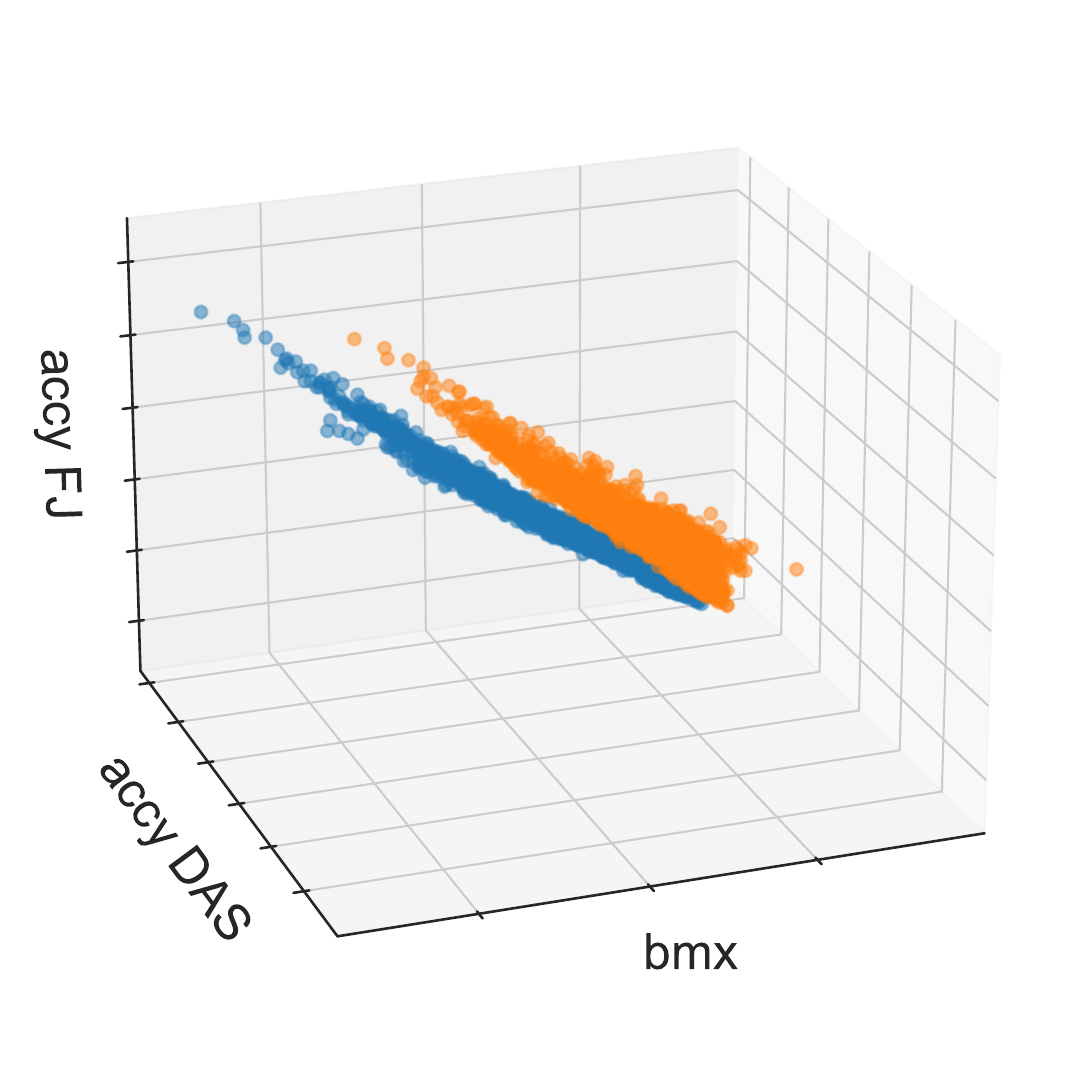}
     \end{subfigure}
     \hfill
    \caption{To the left  a pair plot of the data after using aforementioned standard deviation transform on wells with a tight wellhead housing. For certain combinations the broken and intact cases separate quite well. To the right a 3D plot showing the spread of the data}
    \label{fig:tight_std_pairplot_app}
\end{figure}

\begin{figure}[htbp]
     \centering
     \begin{subfigure}{0.59\textwidth}
        \centering
      \includegraphics[width=\linewidth]{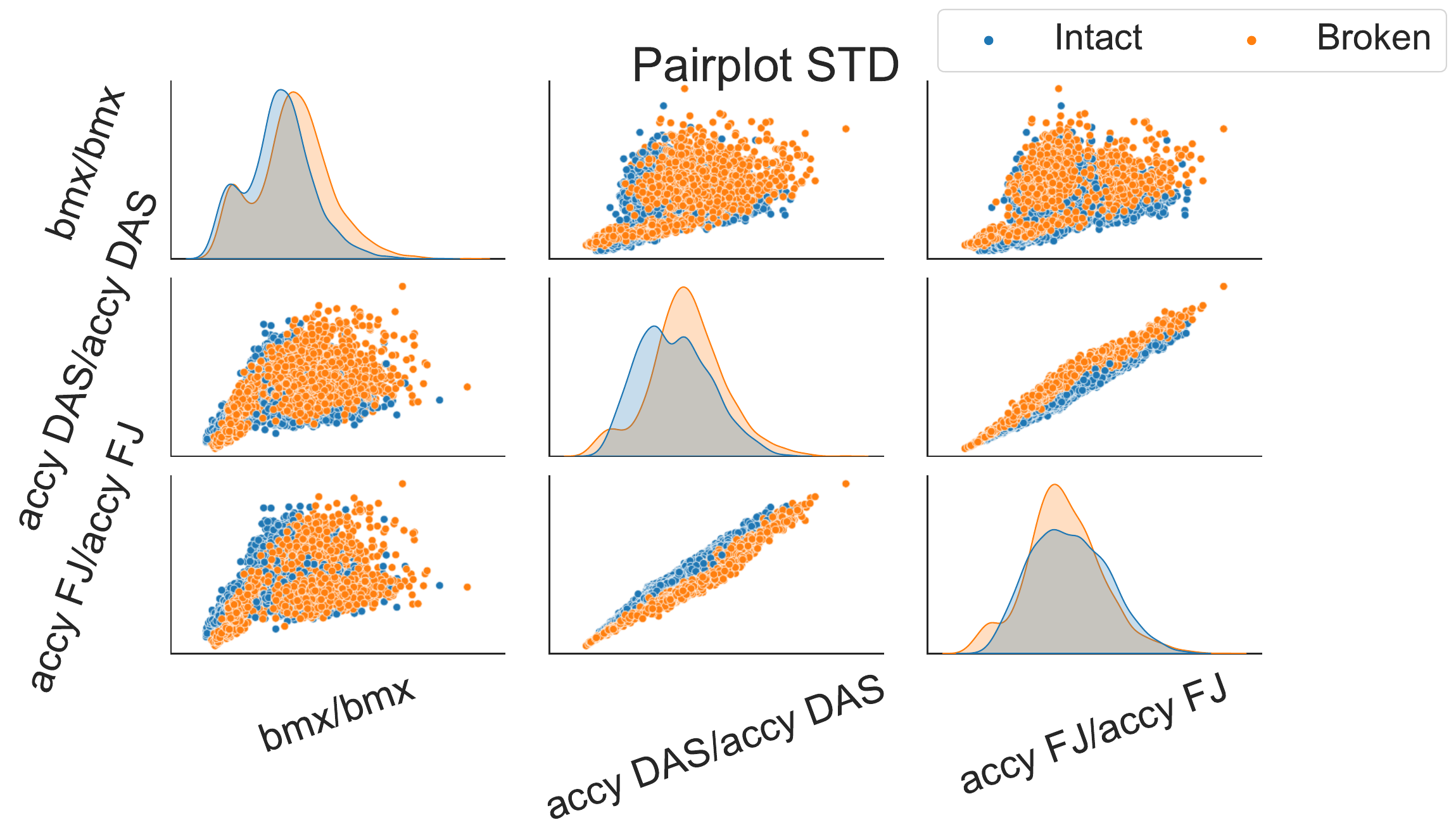}
     \end{subfigure}
     \hfill
     \begin{subfigure}{0.40\textwidth}
        \centering
      \includegraphics[width=\linewidth]{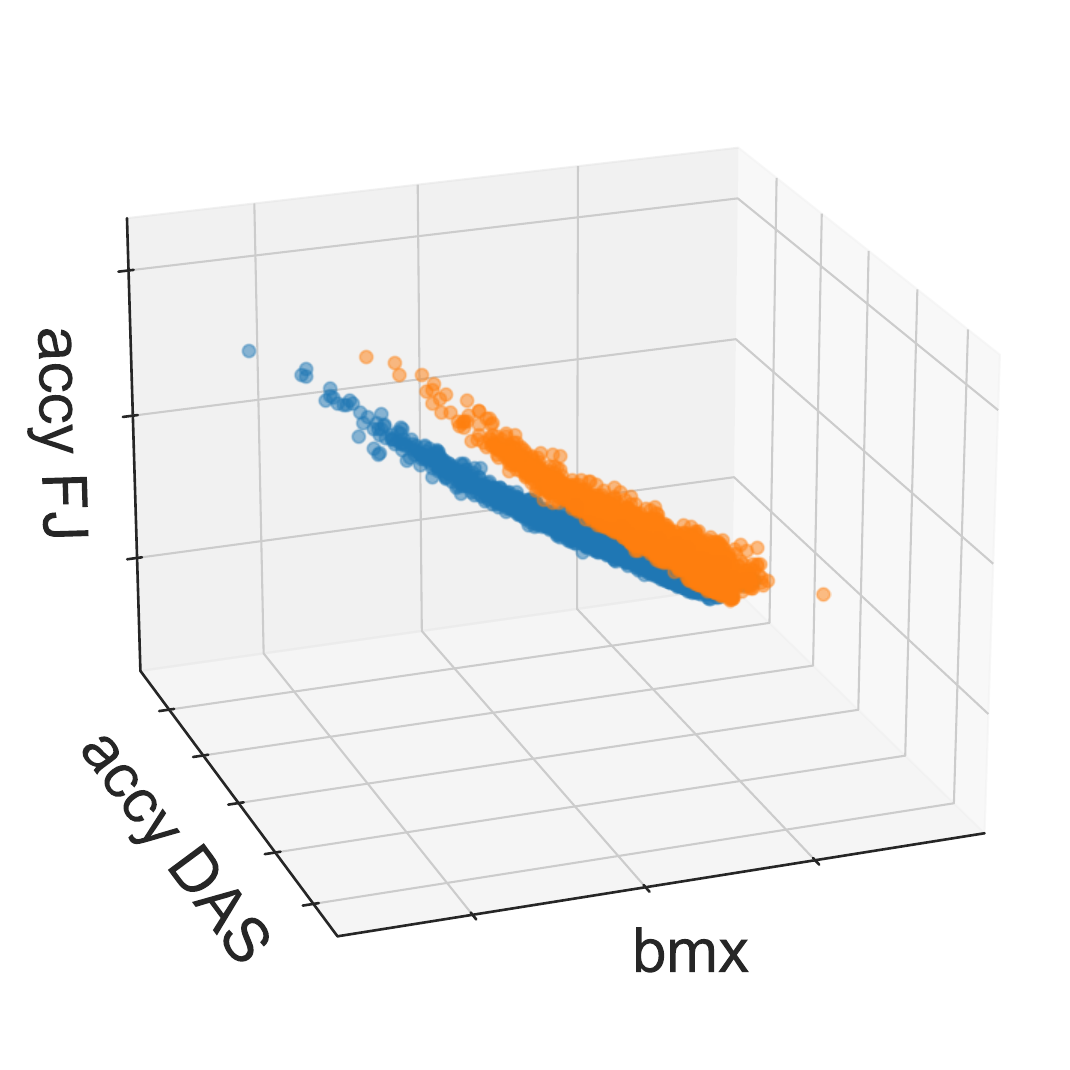}
     \end{subfigure}
     \hfill
    \caption{To the left a pair plot of the data after using aforementioned standard deviation transform on wells with a slack wellhead housing. For certain combinations the broken and intact cases separate quite well. To the right a 3D plot showing the spread of the data}
    \label{fig:slack_std_pairplot_app}
\end{figure}

\begin{figure}[htbp]
    \centering
    \includegraphics[width=\linewidth]{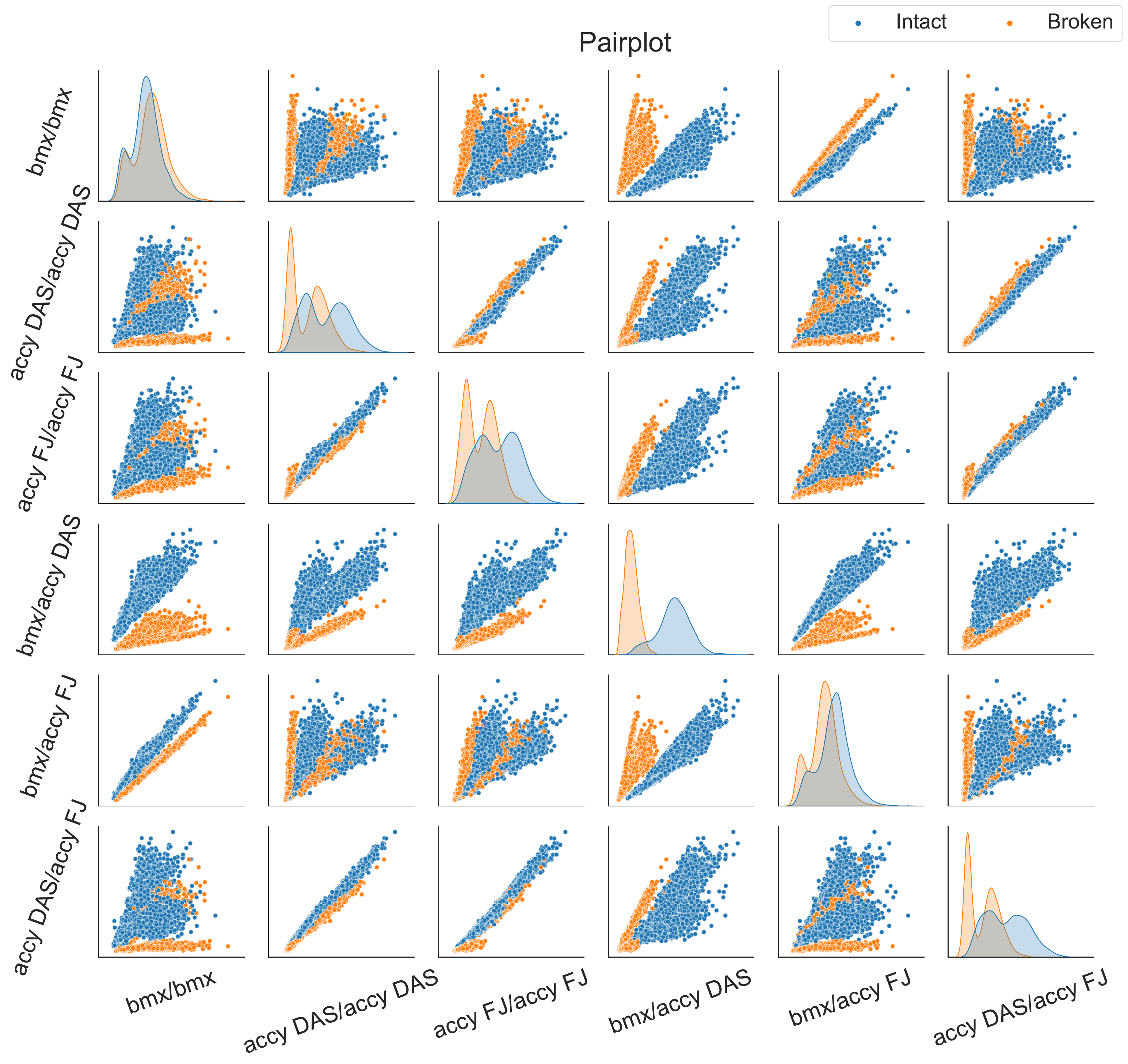}
    \caption{Pair plot of the data after using aforementioned covariance transform on wells with a tight wellhead housing. For certain combinations the broken and intact cases separate quite well.}
    \label{fig:pair_plot_tight}
\end{figure}

\begin{figure}[htbp]
    \centering
    \includegraphics[width=\linewidth]{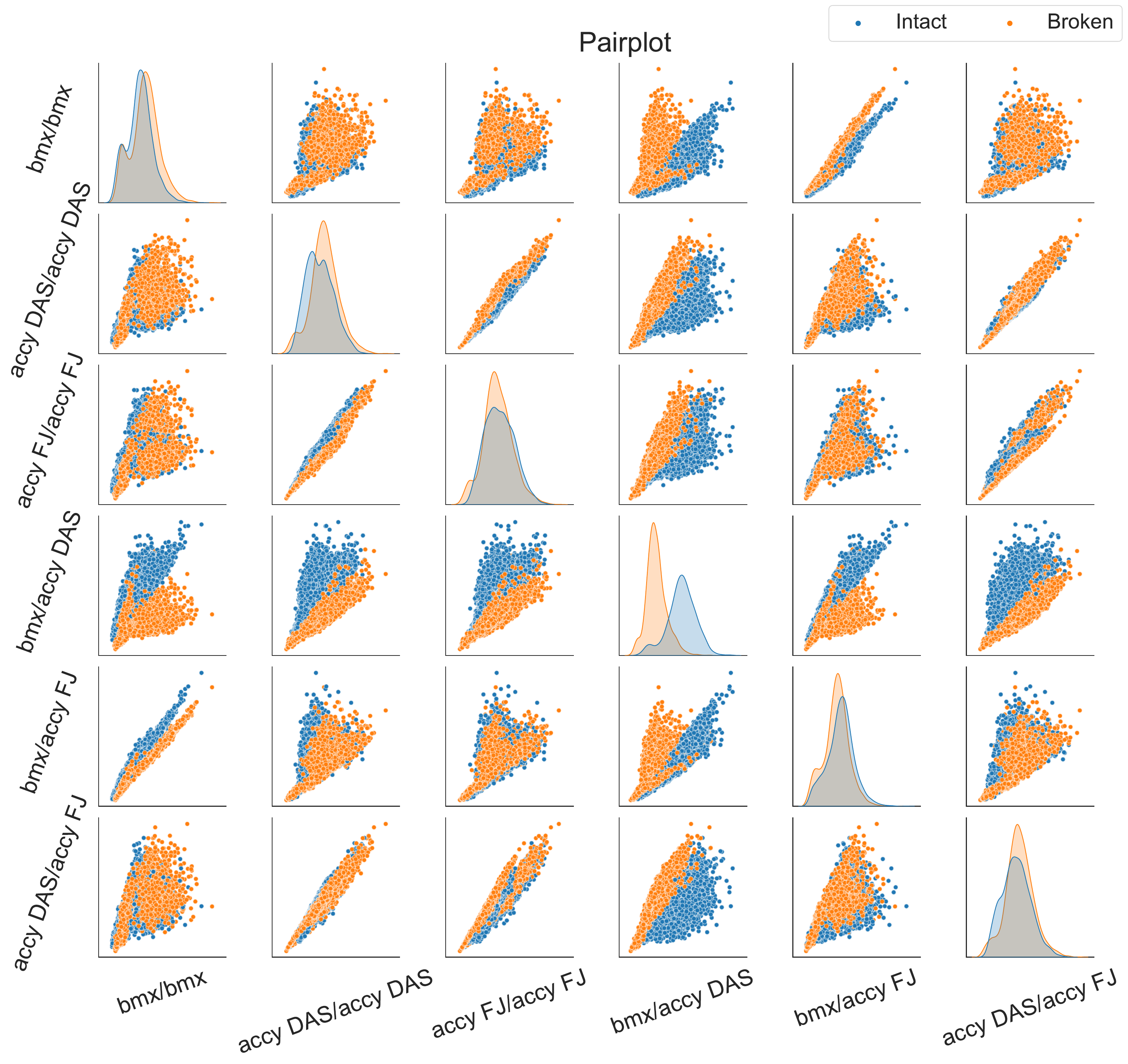}
    \caption{Pair plot of the data after using aforementioned covariance transform on wells with a slack wellhead housing. For certain combinations the broken and intact cases separate quite well.}
    \label{fig:pair_plot_slack_app}
\end{figure}

The specific configuration about the wellhead housing (slack/tight) is of particular importance as one might not be sure about this property before accessing the well. If the wellhead housing is slack the BOP is prone to move more around, which is a similar property to a cracked well. In such case we observe an increased difficulty in classifying on slack data. This becomes apparent when we view the data of the slack and tight WH housing in Figure \ref{fig:tight_std_pairplot_app} to \ref{fig:pair_plot_slack_app}.


Since tight wellhead housing leads to a simpler classification problem than the case with slack, the data set used in the main sections was limited to slack wellhead housing. 

\subsection{Prepocessing the data set}
Whether the time series are passed through a transformation described in Section \ref{sec: dataset} or fed directly to the ML algorithm, they need to be pre-processed to improve performance. 

To standardize the data set’s features to unit scale, i.e., mean equal to 0 and variance equal to 1, we use \texttt{StandardScaler} from \texttt{sklearn.preprocessing}. We may then apply Algorithm \ref{alg:pca_alg} to the standardized training and test set, using \texttt{PCA} from \texttt{sklearn.decomposition}, to reduce the dimensionality.

To train and validate the methods, we divide our data set into a training set and a test set. Typically, these contain 80\% and 20\% of the original data set, respectively. The  machine learning algorithms in this paper makes predictions on the training data and then corrects itself based on the true outputs. Learning stops once the algorithm has achieved an acceptable level of performance on the training set, and the accuracy is measured on the unseen data in the test set. 

\clearpage
\section{Supplementary material for the reproducibility of the experiments: Decision trees}\label{sec:supp_mat_dts}
\textcolor{white}{blanktext}
\newcolumntype{f}{>{\columncolor{mygray}}l}
\begin{table}[h]
\renewcommand{\arraystretch}{1.1}
\begin{tabular}{llffl}
    \hline 
    & & \multicolumn{2}{c}{Pre-pruning} & \text{Post-pruning} \\
    \hline
    $\begin{array}{l}
    \text {Data} \\
    \text {transformation}
    \end{array}$ & $\begin{array}{l}
    \text {Splitting} \\
    \text {criterion}
    \end{array}$ & Hyperparameter  & \text {Range} & \text{$\alpha$}\\
    \hline
  \multirow{6}{*}{STD} & \multirow{3}{*}{Entropy} & \texttt{max\_depth} & $[2, 13]\cap\mathbb{N}$ & \\ 
  & & \texttt{min\_samples\_split} & $[2, 4]\cap\mathbb{N}$ & 0.003\\ 
  & & \texttt{min\_samples\_leaf} & $[1, 2]\cap\mathbb{N}$ & \\ \hhline{~----}
  & \multirow{3}{*}{Gini} & \texttt{max\_depth} & $[2, 13]\cap\mathbb{N}$ & \\ 
  & & \texttt{min\_samples\_split} & $[2, 4]\cap\mathbb{N}$ & 0.002\\ 
  & & \texttt{min\_samples\_leaf} & $[1, 2]\cap\mathbb{N}$ & \\
  \hline
  \multirow{6}{*}{COV} & \multirow{3}{*}{Entropy} & \texttt{max\_depth} & $[2, 5]\cap\mathbb{N}$ & \\ 
  & & \texttt{min\_samples\_split} & $[2, 4]\cap\mathbb{N}$ & 0.01\\ 
  & & \texttt{min\_samples\_leaf} & $[1, 2]\cap\mathbb{N}$ & \\ \hhline{~----}
  & \multirow{3}{*}{Gini} & \texttt{max\_depth} & $[2, 6]\cap\mathbb{N}$ & \\ 
  & & \texttt{min\_samples\_split} & $[2, 4]\cap\mathbb{N}$ & 0.003\\ 
  & & \texttt{min\_samples\_leaf} & $[1, 2]\cap\mathbb{N}$ & \\
  \hline
  \multirow{6}{*}{COV-PCA(4)} & \multirow{3}{*}{Entropy} & \texttt{max\_depth} & $[2, 8]\cap\mathbb{N}$ & \\  
  & & \texttt{min\_samples\_split} & $[2, 4]\cap\mathbb{N}$ & 0.01*\\ 
  & & \texttt{min\_samples\_leaf} & $[1, 2]\cap\mathbb{N}$ & \\ \hhline{~----}
  & \multirow{3}{*}{Gini} & \texttt{max\_depth} & $[2, 8]\cap\mathbb{N}$ & \\ 
  & & \texttt{min\_samples\_split} & $[2, 4]\cap\mathbb{N}$ & 0.003\\ 
  & & \texttt{min\_samples\_leaf} & $[1, 2]\cap\mathbb{N}$ & \\
  \hline
\end{tabular}\caption{\label{tab:dt_params_table}Hyperparameter ranges for the pre-pruning and choice of the $\alpha$ for the post-pruning of the DTs, used to obtain the results reported in Table \ref{tab:dt_table}. \\ * except for the \emph{Noise 50} data set where $\alpha$ = 0.003.}
\end{table}

\section{Supplementary material for the reproducibility of the experiments: Convolutional Neural Networks}\label{sec:supp_mat_cnn}
\textcolor{white}{blanktext}
\begin{table}[htbp]
\centering
\small
{
\begin{tabularx}{1.\textwidth}{ 
   >{\centering\arraybackslash}p{0.35\textwidth} 
   >{\centering\arraybackslash}p{0.35\textwidth} 
   >{\centering\arraybackslash}p{0.2\textwidth}}
\toprule
Hyperparameter   & Range & Distribution \\
\midrule
     activation function & \{Tanh, Swish, Sigmoid, ReLU, LeakyReLU\} & discrete uniform \\
     learning rate & $[1 \cdot 10^{-4} , 1 \cdot 10^{-1}]$ & log uniform \\
     weight decay & $[1 \cdot 10^{-7} , 5 \cdot 10^{-4}]$ & log uniform\\
     batch size & $\{10, 30, 50, 100\}$ & discrete uniform \\
 \bottomrule
\end{tabularx}
}
    \caption{Range of values allowed for each hyperparameter in the experiments with CNNs, with the third column describing how the values were explored using \texttt{Optuna}.}
\label{tab:hyperparams_cnn}
\end{table}

\begin{lstlisting}[language=Python, caption=Architecture of the CNN used in the eperiments of Section \ref{sec: cnn}.]
class cnnseries(nn.Module):
    def __init__(self, act_name='lrelu'):
        super(cnnseries, self).__init__()
        
        torch.manual_seed(1)
        np.random.seed(1)
        random.seed(1)
        
        self.conv1 = torch.nn.Conv1d(in_channels = 6, out_channels = 12, kernel_size = 30, stride=1, padding=0, dilation=1, groups=1, bias=True)
        self.conv2 = torch.nn.Conv1d(in_channels = 12, out_channels = 24, kernel_size = 30, stride=1, padding=0, dilation=1, groups=1, bias=True, padding_mode='zeros', device=None, dtype=None)
        self.conv3 = torch.nn.Conv1d(in_channels = 24, out_channels = 48, kernel_size = 30, stride=1, padding=0, dilation=1, groups=1, bias=True, padding_mode='zeros', device=None, dtype=None)
       
        self.avgpool = torch.nn.AvgPool1d(kernel_size = 15, stride=5, padding=0, ceil_mode=False, count_include_pad=True)

        self.fc2 = nn.Linear(2, 1, bias=True, device=None, dtype=None)
        self.fc1 = nn.Linear(48, 2, bias=True, device=None, dtype=None)        

        self.act_dict = {"tanh":lambda x : torch.tanh(x),
                         "sigmoid":lambda x : torch.sigmoid(x),
                         "swish":lambda x : x*torch.sigmoid(x),
                         "relu":lambda x : torch.relu(x),
                         "lrelu":lambda x : F.leaky_relu(x)}
        self.act = self.act_dict[act_name]

    def forward(self, x):                        
        x = self.act(self.conv1(x))
        x = self.avgpool(x)                       
        x = self.act(self.conv2(x))                
        x = self.avgpool(x)
        x = x.view(x.size(0), -1)
        x = self.act(self.fc1(x))        
        x2 = x
        x = torch.sigmoid(self.fc2(x))
        return x, x2
\end{lstlisting}
\clearpage

\bibliographystyle{AIMS.bst}
\bibliography{230828-Gustad-references}

\medskip
\medskip

\end{document}